%% file: main.tex
\documentclass{article}
\usepackage[OT1]{fontenc}
\usepackage{iclr2027_conference,times}
\input{math_commands.tex}

\usepackage{url}
\usepackage{amsmath,amssymb}
\usepackage{booktabs}
\usepackage{xcolor}
\usepackage{colortbl}
\usepackage{graphicx}
\usepackage{float}
\usepackage{multirow}
\usepackage{placeins}
\usepackage{hyperref}
\definecolor{tablegroup}{HTML}{E7E7E7}
\definecolor{tableours}{HTML}{EAF1F8}
\definecolor{tablegain}{HTML}{087F7B}
\definecolor{tableloss}{HTML}{B45B68}
\definecolor{tableneutral}{HTML}{777777}
\newcommand{\gain}[1]{\textsubscript{\textcolor{tablegain}{(+#1)}}}
\newcommand{\loss}[1]{\textsubscript{\textcolor{tableloss}{(-#1)}}}

\newcommand{\muted}[1]{\textcolor{tableneutral}{#1}}
\newcommand{\tabcite}[1]{{\fontsize{6.2}{6.8}\selectfont\citep{#1}}}
\newcommand{\scoreplain}[1]{%
  \makebox[5.40em][c]{#1}}
\newcommand{\scoregain}[2]{%
  \makebox[2.85em][c]{#1}\makebox[2.55em][l]{\gain{#2}}}
\newcommand{\scoreloss}[2]{%
  \makebox[2.85em][c]{#1}\makebox[2.55em][l]{\loss{#2}}}
\newcommand{\ovplain}[1]{%
  \makebox[4.70em][c]{#1}}
\newcommand{\ovgain}[2]{%
  \makebox[2.55em][c]{#1}\makebox[2.15em][l]{\gain{#2}}}
\newcommand{\ovloss}[2]{%
  \makebox[2.55em][c]{#1}\makebox[2.15em][l]{\loss{#2}}}

\newcommand{\method}{StreamWeave}
\newcommand{\methodcolor}{%
  \textbf{\textup{\textcolor{black}{Stream}%
  \textcolor[HTML]{C95459}{W}%
  \textcolor[HTML]{B47A13}{e}%
  \textcolor[HTML]{39865B}{a}%
  \textcolor[HTML]{347EA9}{v}%
  \textcolor[HTML]{8968A8}{e}}}}
\newcommand{\oursrowshade}[1]{%
  \makebox[0pt][l]{\color{tableours}%
    \rule[-\dp\strutbox]{#1}{\dimexpr\ht\strutbox+\dp\strutbox\relax}}}
\title{\texorpdfstring{\methodcolor}{StreamWeave}: Curating Multi-Timescale Visual Evidence for Streaming Video Understanding}
\author{Hewen Pan\textsuperscript{1,2,*}, Chao Yuan\textsuperscript{2,*}, Jiyang Qi\textsuperscript{2}, Daoxin Zhang\textsuperscript{2}, Cheng Chen\textsuperscript{2}\\
\normalfont\textsuperscript{1}Huazhong University of Science and Technology\\
\normalfont\textsuperscript{2}Xiaohongshu Inc.\\
\normalfont\textsuperscript{*}Equal contribution.}
\newcommand{\ourslabel}{\hspace*{0.9em}\textbf{+\methodcolor{}}}
\iclrfinalcopy

\begin{document}
\maketitle
\lhead{} 
\renewcommand{\headrulewidth}{0pt}

\begin{abstract}
Multimodal large language models have advanced offline video understanding, but streaming video systems must answer questions using only observations available when each question arrives. As the observed prefix grows beyond the model's finite visual context, different questions make different observations relevant. A fixed frame selection cannot reliably preserve all such evidence. Even when past frames are retained, event descriptions written before a question arrives may not reveal which frames contain the visual evidence it seeks. We introduce \methodcolor{}, a training-free agent that curates multi-timescale visual evidence for each question. It builds a coverage-aware visual history linked to source frames and combines a self-curated glance with a broader look-back. The MLLM answers from this context or requests historical frames as needed. Across three streaming video understanding benchmarks and five frozen MLLMs ranging from an 8B dense model to a 397B mixture-of-experts model, \method{} achieves competitive results against published streaming methods. Extensive ablation studies validate the design choices, while budget and latency analyses assess the accuracy--cost trade-off.
\end{abstract} 

\section{Introduction}

Multimodal large language models have advanced offline video understanding,
from open-ended dialogue and audio-visual reasoning to questions requiring
evidence across long recordings
\citep{zhang2023video,maaz2024video,wu2024longvideobench,pan2026ufvideo}.
Continuous video assistants \citep{wang2026streambridge}, time-sensitive
livestream moderation \citep{yew2026dynamic}, and robots acting on ongoing
observations \citep{driess2023palm} need related capabilities while video
is still unfolding.  Streaming video understanding requires each question to
be answered from the prefix observed at its arrival, without relying on
future frames \citep{niu2025ovo,lin2026streamingbench}.  As that prefix
grows, deciding which observations to place in the MLLM's finite visual
context becomes part of the task.

\begin{figure}[!t]
\centering
\includegraphics[width=\linewidth]{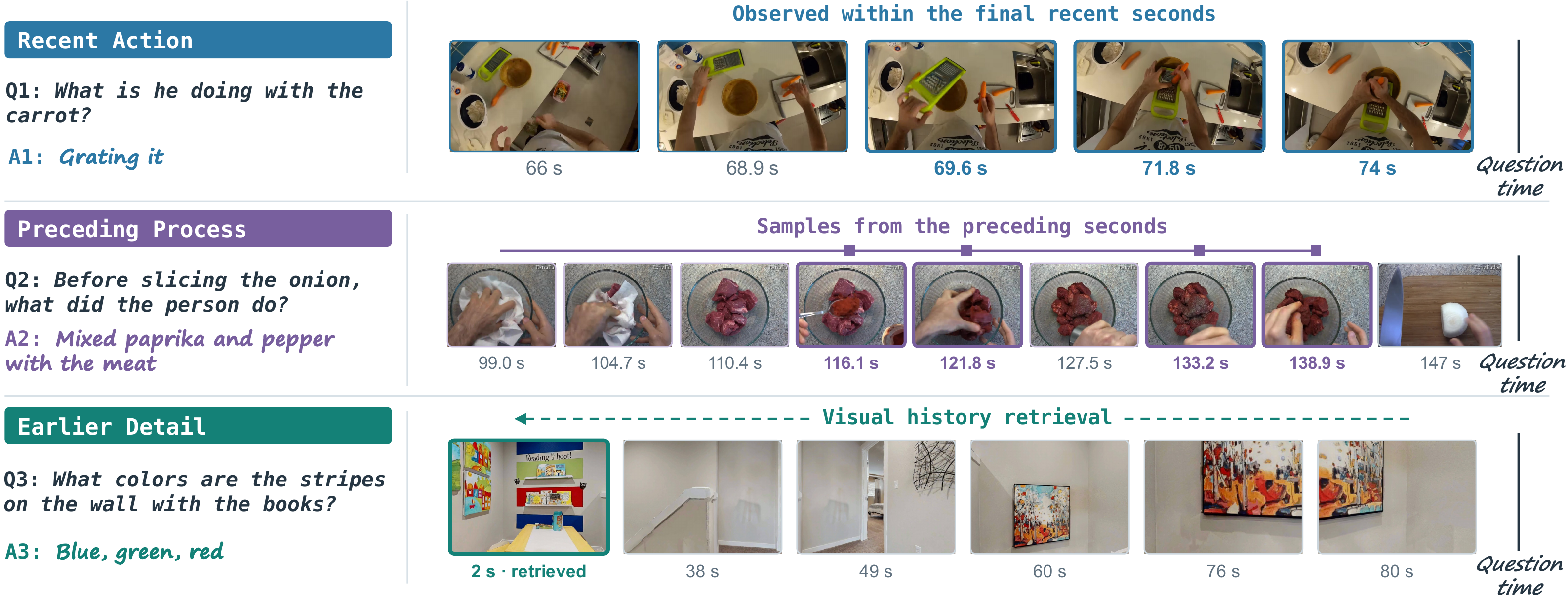}
\caption{Three OVO-Bench examples from separate videos.  A recent action is
resolved near the query frontier (top); observations across a longer interval
show the preceding process (middle); and a briefly observed detail is
recovered from visual history after the question arrives (bottom).  All
displayed frames lie within the observed prefix.}
\label{fig:motivation}
\end{figure}

The consequences of this choice are visible in recent streaming results.
SimpleStream shows that a small recent window can be a strong baseline, while
adding historical context does not necessarily improve real-time perception
\citep{shen2026simple}.  A recent window, however, cannot show every
step of a process that has already passed; spreading a fixed number of samples
over a longer interval increases temporal reach but reduces the density needed
to see short actions \citep{qu2025does}.  Figure~\ref{fig:motivation}
makes the distinction concrete: \emph{Recent Action} depends on continuity
near the question, whereas \emph{Preceding Process} draws on observations
distributed over the preceding activity.  These are different demands on the
same limited input.  Repeated views can spend that input without adding
distinct evidence, so the visual context must balance temporal reach with
detail near the question.

Any bounded selection from the observed video can omit evidence needed by a
later question.  \emph{Earlier Detail} in
Figure~\ref{fig:motivation} illustrates a brief appearance missed by broad
sampling, but historical questions are not limited to isolated details: they
may ask about several occurrences or relate observations made at different
moments.  EventMemAgent and OASIS retain visual evidence from the stream,
but their semantic search relies on event descriptions produced as the video
unfolds \citep{wen2026eventmemagent,liang2026oasis}.  A later question
may hinge on a visual detail that those descriptions did not emphasize.  This
motivates indexing historical observations by their visual content, so the
question can locate relevant records and recover their source frames for
reasoning.

These demands motivate curating multi-timescale visual evidence for changing
questions.  We present \methodcolor{}, a training-free agent that manages both
the observations retained from the stream and the evidence supplied for
reasoning. It keeps current action and preceding activity available while
allowing each question to recover relevant evidence from the observed past.
Its coverage-aware visual history consolidates similar
observations into source-linked records with temporal coverage.  At query time, StreamWeave
assembles a bounded view from a recent glance and a broader look-back,
curating redundant glance frames with the MLLM's visual encoder.  The MLLM
answers from this view or specifies what historical
evidence it needs; query-conditioned matching then returns relevant original
frames to join the same view for final reasoning.  Across five frozen MLLMs,
StreamWeave improves OVO-Bench macro accuracy over model-matched 44-frame
references by 1.7 to 8.4 points and reaches 86.12 on StreamingBench
Real-time.

Our contributions are:
\begin{enumerate}
    \item We introduce \methodcolor{}, a training-free streaming video agent
    that organizes multi-timescale visual evidence for each question within a
    bounded MLLM context.
    \item We develop a coverage-aware visual history with searchable visual
    keys and source-frame links, allowing query-conditioned retrieval to
    complement the glance and look-back with historical evidence.
    \item We evaluate \method{} with five frozen MLLMs on three streaming
    benchmarks, demonstrating consistent OVO-Bench improvements over
    model-matched 44-frame references and examining the roles of glance
    curation, look-back coverage, and history retrieval.
\end{enumerate}

\section{Related Work}

\subsection{Streaming Video Understanding}

Streaming video understanding evaluates a model on the observed prefix rather
than a completed video.  OVO-Bench separates backward, real-time, and forward
capabilities at explicit causal frontiers \citep{niu2025ovo}.
StreamingBench organizes its tasks into Real-Time Visual Understanding,
Omni-Source Understanding, and Contextual Understanding
\citep{lin2026streamingbench}, while OVBench organizes streaming understanding
into perception, memory, verification, and prediction capabilities
\citep{huang2025online}.  LongVideoBench provides a complementary offline
test in which evidence is distributed over long inputs
\citep{wu2024longvideobench}.  Long-video studies also show that sampling
density and the temporal locality of a question affect which visual evidence
is available to a model
\citep{qu2025does,lu2025vited,li2026divide,yuan2025date}.

StreamForest learns persistent event
memory alongside fine-grained current perception, while StreamOPD and
SelectStream also use additional training for streaming understanding
\citep{zeng2026streamforest,wu2026streamopd,ge2026should}.
Training-free work retains the pretrained MLLM while changing which
observations reach it: SimpleStream demonstrates the strength of a recent
window, whereas SAVEMem, FOLIO, and HERMES explore selective or hierarchical
retention \citep{shen2026simple,wu2026semantic,fan2026folio,zhang2026hermes}.
Related work also examines the informational uniqueness of video frames
for compression \citep{yuan2026unicomp}.  Together, these results show why
temporal reach and current-frame detail must be considered jointly.

\subsection{Streaming Video Agents}

Streaming video agents decide what to observe, remember, retrieve, or answer.
StreamBridge couples a memory buffer
with a separate activation model for proactive response, while StreamAgent
plans when and where to observe future task-relevant evidence
\citep{wang2026streambridge,yang2025streamagent}.  StreamRAG introduces
retrieval augmentation for real-time video understanding
\citep{xie2026streamrag}.  EventMemAgent organizes short- and long-term
observations around events and equips the model with adaptive perception tools
\citep{wen2026eventmemagent}.  OASIS combines frame buffers with
description-based access to hierarchical event memory \citep{liang2026oasis}.
Visual Agentic Memory combines visual and temporal retrieval with iterative
inspection of hierarchical memory \citep{li2026visual}; GROVE also develops
agentic access to stored evidence \citep{gong2026grove}. These approaches
motivate examining how retained evidence is organized for each answer.
\method{} couples source-linked coverage records with a self-curated glance
and look-back, adding retrieved original frames to the same bounded context.

\section{Method}

\subsection{Problem setup}

\begin{figure}[t]
\centering
\includegraphics[width=\linewidth]{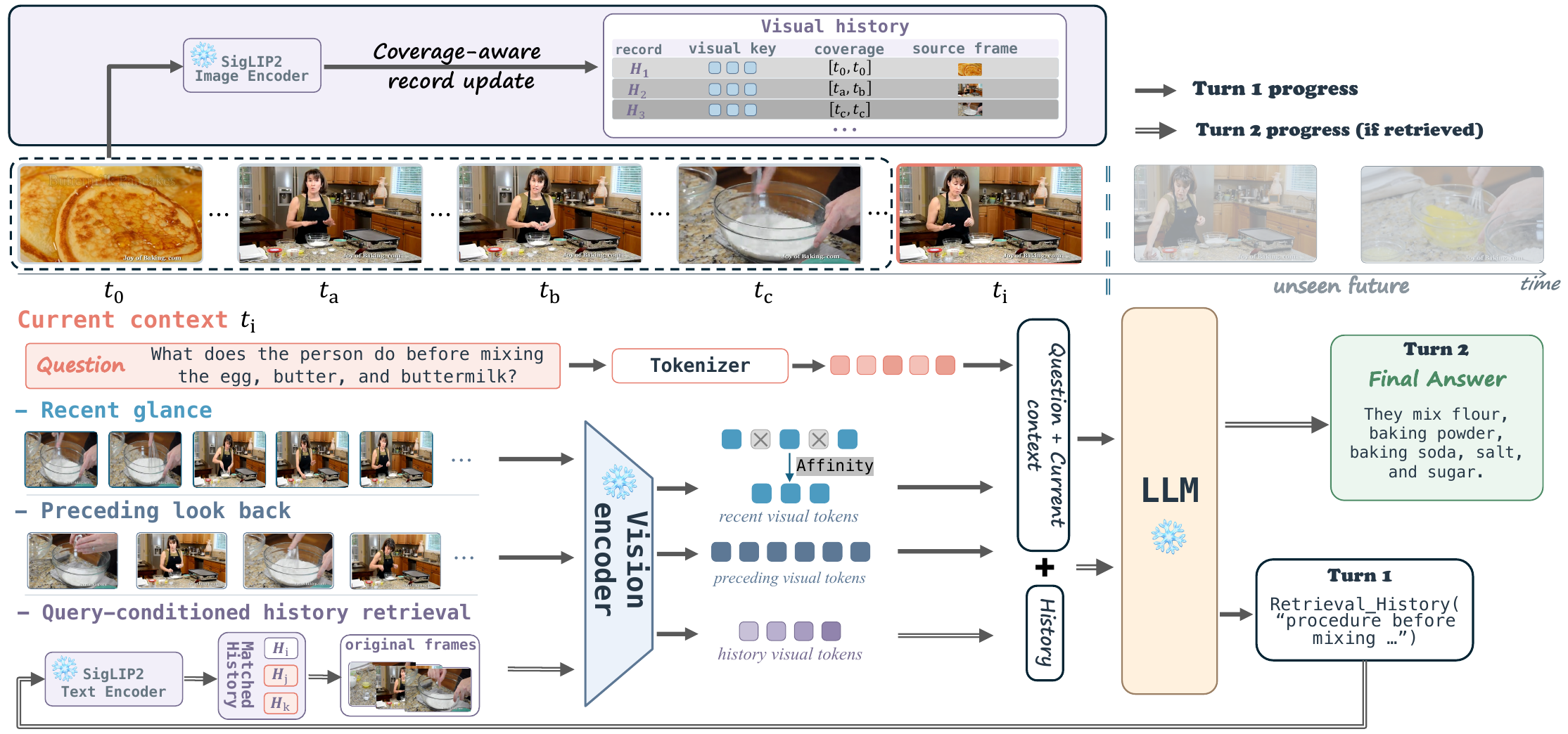}
\caption{Overview of \method{}.  Incoming observations update visual history
records with a visual key, temporal coverage, and a source-frame pointer.
At a query frontier, the glance and look-back form the working context.
A history-retrieval request returns original frames that join the same question and
working context for final reasoning; future frames remain unavailable.}
\label{fig:overview}
\end{figure}

Let $V=(x_t)_{t\geq0}$ be a video stream and let question $q_i$ arrive at
frontier $\tau_i$.  The agent can use only the observed prefix
$V_{\leq\tau_i}=\{x_t\mid t\leq\tau_i\}$.  \method{} maintains a visual
archive $\mathcal A_i$ over this prefix and gives a frozen MLLM $F_\theta$
a working visual context $W_i$ with the question.  The MLLM may answer from
that context or request source frames $H_i$ from the archive for a final call.
At most $B$ images enter a model call, although the archive can grow with the
stream.  Figure~\ref{fig:overview} shows how the current context and
historical evidence reach the same MLLM without exposing future frames.

\subsection{Coverage-aware visual history}
\label{sec:visual-history}

As frames arrive, \method{} builds a visual archive without assigning event
descriptions in advance.  A retained record is
\begin{equation}
    m_j=(h_j,\bar x_j,[a_j,b_j],f_j,p_j),
\end{equation}
where $h_j$ identifies the record, $\bar x_j$ is its representative
observation, and $[a_j,b_j]$ records its temporal coverage.  The visual key
$f_j$ is a frozen SigLIP2 image embedding of that observation
\citep{tschannen2025siglip}; $p_j$ points to the original source frame.
The key makes the record searchable, while the pointer allows the MLLM to
inspect the corresponding pixels after retrieval.

Coverage aggregation updates this archive as frames arrive.  For each
incoming candidate $x_t$, a low-resolution appearance gate checks brightness,
color, and changed-pixel ratios against the latest retained representative.
On a match, the record's coverage
becomes $[a_j,t]$ while its visual key and source-frame pointer stay fixed;
otherwise a new record is appended with coverage $[t,t]$.  Scene changes and
query-time anchors are retained.  Thus repeated observations share a
searchable record, while distinct observations preserve separate source-frame
references.

The appearance gate determines whether to retain a new observation;
query-conditioned visual-key matching later determines which records to
retrieve. Each selected record contributes its representative source frame,
with the coverage interval recording the observations it represents.  Appendix~\ref{sec:protocol-details} gives the
gate's settings.  The eligible archive $\mathcal A_i$ contains only
observations up to $\tau_i$.

\subsection{Weaving multi-timescale visual evidence}

When question $q_i$ arrives at frontier $\tau_i$, the MLLM first receives two
views of the observed stream.  A recent glance retains the action near the
question, while a sparser look-back shows how the scene reached its current
state.  Let $\Delta_g$ and $\Delta_\ell$ be their durations, and let $n_g$
and $n_\ell$ be their frame budgets.  With $\mathcal U(I,k)$ denoting up to
$k$ uniformly selected source frames from interval $I$, we form
\begin{equation}
    R_i=\mathcal U([\tau_i-\Delta_g,\tau_i],n_g),\qquad
    L_i=\mathcal U([\tau_i-\Delta_g-\Delta_\ell,\tau_i-\Delta_g),n_\ell).
\end{equation}
Here $R_i$ contains glance candidates and $L_i$ contains look-back frames.
Both are available in the first model call.  The look-back remains unchanged;
the glance is curated to avoid spending visual input on nearly identical
adjacent observations.

The MLLM's visual encoder represents each glance candidate $r_j$ as a pooled
feature $v_j$.  We score an interior frame by its similarity to both
neighbors in the original glance:
\begin{equation}
\begin{aligned}
    v_j&=\operatorname{L2Norm}\!\left(
       \operatorname{MeanPool}(E_m^{\mathrm{vis}}(r_j))\right),\\
    \rho_j&=\min(v_{j-1}^{\top}v_j,\;v_j^{\top}v_{j+1}).
\end{aligned}
\end{equation}
We protect the first and final three glance positions, then greedily remove
nonadjacent interior frames with $\rho_j\geq\gamma$ in descending score
order, retaining at least $\min(|R_i|,9)$ glance frames.  The retained
sequence $R_i^\star$ and look-back frames $L_i$ form
$W_i=\operatorname{Pack}(L_i,R_i^\star)$, preserving source timestamps and
temporal order.

The frozen MLLM first reasons over $(q_i,W_i)$.  It can answer from this
context or call the history-retrieval tool to request missing evidence.  For questions
explicitly about earlier events, the model may reconsider a direct response
once before it is finalized.  A history request $u_i$ expresses the
information sought as text $z_i$ and may specify temporal, evidence-type,
scope, and frame-budget constraints.  The SigLIP2 text encoder maps $z_i$
into the embedding space of the visual keys $f_j$ stored in
Section~\ref{sec:visual-history}:
\begin{equation}
    g_i=\operatorname{L2Norm}(E_{\mathrm S}^{\mathrm{text}}(z_i)),
    \qquad s_{ij}=g_i^\top f_j,\quad m_j\in\mathcal A_i.
\end{equation}
The tool ranks eligible records by $s_{ij}$, applies the request constraints,
and excludes source frames already present in $W_i$.  Several records can be
selected from different parts of the observed prefix.  Each contributes one
original source frame, and the returned frames are ordered by timestamp:
\begin{equation}
\begin{aligned}
    J_i&=\operatorname{SelectRecords}
       \bigl(u_i,\{(m_j,s_{ij}):m_j\in\mathcal A_i\}\bigr),
       \qquad |J_i|\leq k_i,\\
    H_i&=\operatorname{SortTime}
       \bigl((\operatorname{Load}(p_j))_{j\in J_i}\bigr).
\end{aligned}
\end{equation}

When history is retrieved, the same MLLM receives the original question and
the complete first-call context together with the retrieval trace and
recalled source frames.  It produces the final answer from
$C_i=\operatorname{Pack}(W_i,H_i)$, where $n_{\mathrm{img}}(C_i)\leq B$.
The $\operatorname{Pack}$ operation preserves group labels, timestamps, and
within-group order; recalled frames follow the first-call context.  History
can therefore contribute evidence from anywhere in the observed prefix
without becoming another fixed sampling window.

The complementary roles of these views can be expressed as an evidence
decomposition.  Let $Y$ denote the answer, $Q$ the question, and
$R^\star$, $L$, and $H$ the corresponding visual evidence over queries.
The chain rule gives
\begin{equation}
\begin{aligned}
I(Y;R^\star,L,H\mid Q)
={}&I(Y;R^\star\mid Q)\\
&+I(Y;L\mid Q,R^\star)\\
&+I(Y;H\mid Q,R^\star,L).
\end{aligned}
\end{equation}
The conditional terms describe what each view adds beyond the evidence
already present: current detail in the glance, preceding context in the
look-back, and historical evidence recovered for the question. Their
contributions can therefore vary across questions and tasks.

\section{Experiments}
\label{sec:experiments}

\subsection{Experimental setup}

\paragraph{Benchmarks.}
We evaluate on three streaming benchmarks.  OVO-Bench has 3,035 questions in
Backward, Real-time, and Forward tracks; its overall score is the macro average
of the three track accuracies \citep{niu2025ovo}.  We use all 2,500 questions
in the StreamingBench Real-time split and report both its official accuracy and
ten task scores \citep{lin2026streamingbench}.  OVBench comprises six task
families and 16 subtasks \citep{huang2025online}.  We show the six families
in the main table and the complete subtask results in
Appendix~\ref{sec:ovbench-detailed}; the
official overall score averages the 16 subtask accuracies, not the six family
means.  A separate LongVideoBench transfer study appears in
Appendix~\ref{sec:lvb-siglip-transfer}
\citep{wu2024longvideobench}.

\vspace{-0.5cm}

\paragraph{Models and protocol.}
We use five frozen MLLMs spanning dense and mixture-of-experts architectures:
Qwen3-VL-8B-Instruct, Qwen3.5-9B,
Qwen3-VL-32B-Instruct, Qwen3.6-27B, and the FP8 checkpoint of
Qwen3.5-397B-A17B.  We set the glance duration $\Delta_g$ to 8 s with
$n_g=12$ candidates and the look-back duration $\Delta_\ell$ to 40 s with
$n_\ell=8$ frames.  The MLLM's visual encoder curates the glance candidates;
SigLIP2 provides the keys for historical search
\citep{tschannen2025siglip}.  Coverage aggregation uses a gate threshold of
0.85; glance curation uses a cosine-similarity threshold of $\gamma=0.995$.
A retrieval can return up to 24 original
frames, giving at most 44 images in the final visual context.  The same
configuration is used across the five MLLMs and three benchmarks without 
benchmark-specific tuning, allowing the same evidence-curation policy to be
evaluated across model scales and architectures.

For evaluation throughput, we cache query-independent visual features and
history updates, then restrict the archive to each question's observed prefix.
The MLLM always receives original frames rather than index features.
\method{} uses temperature 0 and at most 256 output tokens. Sampling and
answering protocols, including the Qwen3.6-27B uniform baselines, are specified
in Appendix~\ref{sec:protocol-details}. Qwen3.6-27B component, selection, and
efficiency studies use the common candidate-pool protocol in
Appendix~\ref{sec:additional-studies}.

For each MLLM, the base uniformly samples 44 frames from the observed prefix.
Tables~\ref{tab:ovo-main}--\ref{tab:ovbench-main} retain the published methods'
frame protocols and use dashes for unreported task scores.  Paired OVO-Bench
intervals use 10,000 source-video cluster-bootstrap replicates.

\subsection{OVO-Bench}

OVO-Bench exposes different evidence demands across Backward, Real-time,
and Forward questions.  Table~\ref{tab:ovo-main} compares each
MLLM with its model-matched 44-frame base and with published streaming
systems. Across all five MLLMs, \method{} raises the three-track macro
average by 1.7 to 8.4 points; every model improves on both Backward and
Real-time.
\vspace{-8pt}

\begin{table}[H]
\caption{OVO-Bench accuracy (\%). \#Frames follows each source's protocol;
counts denote per-query visual context and fps denotes stream sampling.
Colored subscripts report percentage-point changes from the corresponding
44-frame reference;
bold and underlined scores mark the best and second best.}
\label{tab:ovo-main}
\vspace{4pt}
\centering
{\fontsize{8.0}{8.9}\selectfont
\setlength{\tabcolsep}{0.55pt}
\renewcommand{\arraystretch}{0.98}
\begin{tabular*}{\textwidth}{@{\extracolsep{\fill}}lcc*{4}{c}@{}}
\toprule
Method & Params & \#Frames & Backward & Real-time & Forward & Avg. \\
\midrule
\rowcolor{tablegroup}\multicolumn{7}{l}{\muted{\textit{Training-based methods}}} \\
\muted{StreamForest\,\tabcite{zeng2026streamforest}} & \muted{7B} & \muted{1 fps}
    & \muted{\scoreplain{52.00}} & \muted{\scoreplain{61.20}}
    & \muted{\scoreplain{53.50}} & \muted{\scoreplain{55.60}} \\
\muted{EventMemAgent\,\tabcite{wen2026eventmemagent}} & \muted{8B} & \muted{$\leq 32$}
    & \muted{\scoreplain{58.03}} & \muted{\scoreplain{68.29}}
    & \muted{\scoreplain{55.92}} & \muted{\scoreplain{60.75}} \\
\muted{SelectStream\,\tabcite{ge2026should}} & \muted{8B} & \muted{1 fps}
    & \muted{\scoreplain{62.20}} & \muted{\scoreplain{\textbf{82.76}}}
    & \muted{\scoreplain{56.13}} & \muted{\scoreplain{67.03}} \\
\midrule
\rowcolor{tablegroup}\multicolumn{7}{l}{\textit{Training-free methods}} \\
SimpleStream\,\tabcite{shen2026simple} & 8B & 4
    & \scoreplain{54.00} & \scoreplain{81.40} & \scoreplain{--} & \scoreplain{--} \\
HERMES\,\tabcite{zhang2026hermes} & 7B & 1 fps
    & \scoreplain{49.40} & \scoreplain{69.00} & \scoreplain{--} & \scoreplain{--} \\
OASIS\,\tabcite{liang2026oasis} & 8B & --
    & \scoreplain{57.21} & \scoreplain{78.14} & \scoreplain{--} & \scoreplain{--} \\
FOLIO\,\tabcite{fan2026folio} & 8B & --
    & \scoreplain{69.10} & \scoreplain{\underline{82.00}}
    & \scoreplain{--} & \scoreplain{--} \\
GROVE\,\tabcite{gong2026grove} & Mixed & 0.5 fps
    & \scoreplain{\underline{69.90}} & \scoreplain{76.20}
    & \scoreplain{56.50} & \scoreplain{67.50} \\
FluxMem\,\tabcite{xie2026fluxmem} & 7B & 1 fps
    & \scoreplain{--} & \scoreplain{67.20}
    & \scoreplain{--} & \scoreplain{53.30} \\
SAVEMem\,\tabcite{wu2026semantic} & 7B & 1 fps
    & \scoreplain{50.44} & \scoreplain{74.93}
    & \scoreplain{--} & \scoreplain{--} \\
\midrule
\textit{Qwen3-VL-8B} & 8B & 44
    & \scoreplain{51.94} & \scoreplain{68.25} & \scoreplain{54.50} & \scoreplain{58.23} \\
\oursrowshade{\textwidth}\ourslabel & 8B & $\leq 44$
    & \scoregain{58.89}{7.0} & \scoregain{74.40}{6.1}
    & \scoreloss{53.69}{0.8} & \scoregain{62.33}{4.1} \\
\addlinespace[1pt]
\textit{Qwen3.5-9B} & 9B & 44
    & \scoreplain{51.37} & \scoreplain{71.66} & \scoreplain{54.34} & \scoreplain{59.12} \\
\oursrowshade{\textwidth}\ourslabel & 9B & $\leq 44$
    & \scoregain{60.33}{9.0} & \scoregain{76.43}{4.8}
    & \scoregain{55.83}{1.5} & \scoregain{64.19}{5.1} \\
\addlinespace[1pt]
\textit{Qwen3-VL-32B} & 32B & 44
    & \scoreplain{60.77} & \scoreplain{72.28} & \scoreplain{54.25} & \scoreplain{62.43} \\
\oursrowshade{\textwidth}\ourslabel & 32B & $\leq 44$
    & \scoregain{63.00}{2.2} & \scoregain{77.46}{5.2}
    & \scoreloss{51.80}{2.4} & \scoregain{64.09}{1.7} \\
\addlinespace[1pt]
\textit{Qwen3.6-27B} & 27B & 44
    & \scoreplain{58.22} & \scoreplain{74.81} & \scoreplain{49.77} & \scoreplain{60.94} \\
\oursrowshade{\textwidth}\ourslabel & 27B & $\leq 44$
    & \scoregain{65.64}{7.4} & \scoregain{80.34}{5.5}
    & \scoregain{\textbf{62.00}}{12.2} & \scoregain{\underline{69.32}}{8.4} \\
\addlinespace[1pt]
\textit{Qwen3.5-397B-A17B} & 397B-A17B-FP8 & 44
    & \scoreplain{60.96} & \scoreplain{73.70}
    & \scoreplain{\underline{59.90}} & \scoreplain{64.85} \\
\oursrowshade{\textwidth}\ourslabel & 397B-A17B-FP8 & $\leq 44$
    & \scoregain{\textbf{72.00}}{11.0} & \scoregain{81.10}{7.4}
    & \scoreloss{57.11}{2.8} & \scoregain{\textbf{70.07}}{5.2} \\
\bottomrule
\end{tabular*}}
\end{table}
\vspace{-8pt}

Forward changes vary by model: Qwen3.6-27B gains 12.2 points and
Qwen3.5-9B gains 1.5 points, while the other three decrease by 0.8 to
2.8 points. Qwen3.5-397B-A17B achieves the highest macro score, 70.07,
and Qwen3.6-27B achieves the largest paired gain, reaching 69.32. The
joint Backward and Real-time gains raise the complete three-track average
across all models, while Forward remains the main source of variation.
GROVE's Mixed label denotes its Qwen3.5-35B-A3B perception model,
GPT-4.1-mini memory consolidation, and GPT-5-mini reasoning
\citep{gong2026grove}. FluxMem reports an overall score for all three
OVO-Bench tracks, whereas SAVEMem reports only Backward and Real-time;
its two-track average is not entered in the three-track Avg. column.
Appendix~\ref{sec:detailed-results} reports the 12-task breakdown.
\vspace{-8pt}

\subsection{StreamingBench Real-time}

StreamingBench Real-time tests whether broader temporal reach comes at the
cost of perception at the query frontier.  With the same
configuration, all five MLLMs improve over their model-matched bases in
Table~\ref{tab:streaming-main}.  Qwen3.5-397B-A17B reaches 86.12, and
Qwen3.6-27B reaches 85.60.  The paired gains range from 0.9 points for the
27B model to 5.0 points for the 397B model; the 8B model also gains 3.8
points. The gains therefore do not simply increase with model size.

Clips Summarize and Spatial Understanding
improve in every paired comparison, whereas Counting decreases in every
one.  For Qwen3-VL-8B, Clips Summarize rises from 80.8 to 92.7 and
Counting falls from 57.0 to 47.1; the same contrast appears in the
397B model, where these scores change from 89.3 to 93.7 and from 53.9 to
48.7, respectively. Appendix~\ref{sec:additional-counting} examines Counting
through separate curation, aggregation, and retrieval ablations.
Qwen3.8-Omni-Flash reports an overall
score of 80.80 without a public ten-task breakdown
\citep{qwen38omniflash}.

\subsection{OVBench}

Real-time gains alone do not show whether the same visual context helps with
other temporal demands.

\begin{table}[t]
\caption{StreamingBench Real-time accuracy (\%). Avg. is the official score
over 2,500 questions; task scores have one decimal. Colored Avg. subscripts
compare with each 44-frame reference. Bold and underline mark the highest and
second-highest reported scores in each column; ties share a mark. Task names
appear in Appendix~\ref{sec:protocol-details}.}
\label{tab:streaming-main}
\vspace{4pt}
\centering
{\fontsize{8.2}{9.2}\selectfont
\setlength{\tabcolsep}{1.8pt}
\renewcommand{\arraystretch}{1.08}
\renewcommand{\scoreplain}[1]{\makebox[4.8em][c]{#1}}
\renewcommand{\scoregain}[2]{\makebox[2.6em][c]{#1}\makebox[2.2em][l]{\gain{#2}}}
\resizebox{\textwidth}{!}{%
\begin{tabular}{@{}lcc*{11}{c}@{}}
\toprule
\multirow{2}{*}{Method} & \multirow{2}{*}{Params} &
\multirow{2}{*}{\#Frames} & \multicolumn{10}{c}{Real-time tasks} &
\multirow{2}{*}{Avg.} \\
\cmidrule(lr){4-13}
& & & OP & CR & CS & ATP & EU & TR & PR & SU & ACP & CT & \\
\midrule
\rowcolor{tablegroup}\multicolumn{14}{l}{\muted{\textit{Training-based methods}}} \\
\muted{StreamForest} & \muted{7B} & \muted{1 fps}
    & \muted{83.1} & \muted{82.8} & \muted{82.7} & \muted{84.3} & \muted{77.5}
    & \muted{78.2} & \muted{76.8} & \muted{69.1} & \muted{75.6} & \muted{54.4} & \muted{\scoreplain{77.26}} \\
\muted{EventMemAgent} & \muted{8B} & \muted{$\leq32$}
    & \muted{83.9} & \muted{76.6} & \muted{87.7} & \muted{85.3} & \muted{77.0}
    & \muted{79.4} & \muted{77.8} & \muted{68.3} & \muted{72.2} & \muted{48.7} & \muted{\scoreplain{77.00}} \\
\muted{StreamOPD} & \muted{4B} & \muted{4}
    & \muted{--} & \muted{--} & \muted{--} & \muted{--} & \muted{--}
    & \muted{--} & \muted{--} & \muted{--} & \muted{--} & \muted{--} & \muted{\scoreplain{84.55}} \\
\muted{SelectStream} & \muted{8B} & \muted{1 fps}
    & \muted{86.3} & \muted{87.4} & \muted{86.0} & \muted{87.1} & \muted{\textbf{84.7}}
    & \muted{84.6} & \muted{\underline{85.8}} & \muted{75.0} & \muted{78.2} & \muted{\textbf{71.6}} & \muted{\scoreplain{82.67}} \\
\midrule
\rowcolor{tablegroup}\multicolumn{14}{l}{\textit{Training-free methods}} \\
SimpleStream & 8B & 4
    & -- & -- & -- & -- & --
    & -- & -- & -- & -- & -- & \scoreplain{80.59} \\
HERMES & 7B & 1 fps
    & 83.7 & 81.3 & 88.0 & 87.5 & 76.7
    & 86.6 & 82.4 & 76.0 & 73.9 & 46.6 & \scoreplain{79.44} \\
OASIS & 8B & --
    & -- & -- & -- & -- & --
    & -- & -- & -- & -- & -- & \scoreplain{78.22} \\
FOLIO & 8B & --
    & -- & -- & -- & -- & --
    & -- & -- & -- & -- & -- & \scoreplain{81.20} \\
GROVE & Mixed & 0.5 fps
    & -- & -- & -- & -- & --
    & -- & -- & -- & -- & -- & \scoreplain{77.00} \\
FluxMem & 7B & 1 fps
    & 80.2 & 81.1 & 81.4 & 85.3 & 78.0
    & 83.8 & 80.6 & 65.9 & 69.6 & 52.1 & \scoreplain{76.40} \\
SAVEMem & 7B & 1 fps
    & -- & -- & -- & -- & --
    & -- & -- & -- & -- & -- & \scoreplain{76.00} \\
\midrule
\rowcolor{tablegroup}\multicolumn{14}{l}{\textit{Off-the-shelf MLLMs}} \\
Qwen3.8-Omni-Flash & -- & --
    & -- & -- & -- & -- & --
    & -- & -- & -- & -- & -- & \scoreplain{80.80} \\
\midrule
\textit{Qwen3-VL} & 8B & 44
    & 81.2 & 71.1 & 80.8 & 84.0 & 74.5
    & 81.0 & 81.5 & 65.5 & 73.9 & 57.0 & \scoreplain{76.08} \\
\oursrowshade{\textwidth}\ourslabel & 8B & $\leq44$
    & 84.5 & 74.2 & 92.7 & 86.9 & 75.8
    & 84.4 & 80.6 & 72.0 & 80.2 & 47.1 & \scoregain{79.84}{3.8} \\
\addlinespace[0.6pt]
\textit{Qwen3.5} & 9B & 44
    & 84.7 & \underline{88.3} & 88.6 & 89.2 & 77.0
    & 86.6 & 80.6 & 76.0 & 80.5 & 57.5 & \scoreplain{81.96} \\
\oursrowshade{\textwidth}\ourslabel & 9B & $\leq44$
    & 88.8 & 86.7 & 91.2 & 90.8 & 77.6
    & 86.9 & 83.3 & \textbf{82.1} & 81.9 & 47.1 & \scoregain{83.20}{1.2} \\
\addlinespace[0.6pt]
\textit{Qwen3-VL} & 32B & 44
    & 82.3 & \underline{88.3} & 87.1 & 85.6 & 78.3
    & 84.1 & 83.3 & 69.9 & 76.8 & 54.9 & \scoreplain{79.52} \\
\oursrowshade{\textwidth}\ourslabel & 32B & $\leq44$
    & 85.6 & 83.6 & \underline{93.1} & 87.3 & 80.8
    & 86.3 & 85.2 & 76.8 & 83.6 & 49.2 & \scoregain{82.44}{2.9} \\
\addlinespace[0.6pt]
\textit{Qwen3.6} & 27B & 44
    & 87.5 & \textbf{92.2} & 89.6 & \underline{91.5} & 82.0
    & \underline{89.4} & 84.3 & 76.0 & 82.7 & \underline{64.8} & \scoreplain{84.68} \\
\oursrowshade{\textwidth}\ourslabel & 27B & $\leq44$
    & \underline{91.0} & \underline{88.3} & 92.7 & \textbf{92.8} & 82.6
    & \underline{89.4} & 84.3 & \underline{81.7} & \textbf{85.8} & 51.8 & \scoregain{\underline{85.60}}{0.9} \\
\addlinespace[0.6pt]
\textit{Qwen3.5} & {\fontsize{6.0}{6.8}\selectfont \textbf{397B-A17B-FP8}} & 44
    & 87.2 & 66.4 & 89.3 & 87.3 & 80.8
    & 85.4 & 84.3 & 74.4 & 82.4 & 53.9 & \scoreplain{81.12} \\
\oursrowshade{\textwidth}\ourslabel & {\fontsize{6.0}{6.8}\selectfont \textbf{397B-A17B-FP8}} & $\leq44$
    & \textbf{92.4} & \underline{88.3} & \textbf{93.7} & \textbf{92.8} & \underline{84.5}
    & \textbf{91.3} & \textbf{86.1} & \textbf{82.1} & \underline{85.5} & 48.7 & \scoregain{\textbf{86.12}}{5.0} \\
\bottomrule
\end{tabular}}}
\end{table}
\FloatBarrier

OVBench provides a complementary test through 16 subtasks organized into six
families. Table~\ref{tab:ovbench-main} reports the family means and the
official 16-subtask average.

\begin{table}[h]
\caption{OVBench accuracy (\%). The six task-family columns average their
constituent subtasks; Avg. is the official mean of all 16 subtask scores.
\#Frames follows each source's protocol. Colored Avg. subscripts compare
with each 44-frame reference; bold and underline mark the highest and
second-highest reported scores in each column.}
\label{tab:ovbench-main}
\vspace{2pt}
\centering
{\fontsize{8.0}{8.9}\selectfont
\setlength{\tabcolsep}{0.8pt}
\renewcommand{\arraystretch}{0.98}
\begin{tabular*}{\textwidth}{@{\extracolsep{\fill}}lcc*{7}{c}@{}}
\toprule
Method & Params & \#Frames & FP & THV & PM & SP & STP & TP & Avg. \\
\midrule
\rowcolor{tablegroup}\multicolumn{10}{l}{\muted{\textit{Training-based methods}}} \\
\muted{VideoChat-Online\,\tabcite{huang2025online}} & \muted{4B} & \muted{2 fps}
    & \muted{46.8} & \muted{61.4} & \muted{55.7}
    & \muted{39.5} & \muted{68.6} & \muted{56.9}
    & \muted{\ovplain{54.90}} \\
\muted{StreamForest\,\tabcite{zeng2026streamforest}} & \muted{7B} & \muted{1 fps}
    & \muted{--} & \muted{--} & \muted{--}
    & \muted{--} & \muted{--} & \muted{--}
    & \muted{\ovplain{60.50}} \\
\midrule
\rowcolor{tablegroup}\multicolumn{10}{l}{\textit{Off-the-shelf MLLMs}} \\
Gemini-1.5-Flash\,\tabcite{team2024gemini} & -- & $\leq64$
    & 49.0 & 52.6 & 49.0 & 34.9 & 68.5 & 50.8 & \ovplain{50.70} \\
LLaVA-OV-7B\,\tabcite{li2024llava} & 7B & $\leq64$
    & 55.5 & 51.7 & 49.4 & 28.6 & 68.1 & 43.0 & \ovplain{49.50} \\
Qwen2-VL-7B\,\tabcite{wang2024qwen2} & 7B & $\leq64$
    & 49.5 & 52.5 & 57.2 & 31.9 & \underline{73.2} & 35.9 & \ovplain{49.70} \\
\midrule
\textit{Qwen3-VL-8B} & 8B & 44
    & 50.2 & 60.8 & 53.2 & 42.1 & 71.5 & 54.3 & \ovplain{55.14} \\
\oursrowshade{\textwidth}\ourslabel & 8B & $\leq44$
    & 54.1 & 59.9 & 52.3 & 40.6 & 70.4 & 53.6 & \ovloss{55.10}{0.04} \\
\addlinespace[1pt]
\textit{Qwen3.5-9B} & 9B & 44
    & 51.3 & 55.6 & 60.3 & 42.1 & 64.3 & 56.2 & \ovplain{55.18} \\
\oursrowshade{\textwidth}\ourslabel & 9B & $\leq44$
    & 56.4 & 62.6 & 53.7 & \textbf{50.5} & 67.4 & 59.3 & \ovgain{58.23}{3.05} \\
\addlinespace[1pt]
\textit{Qwen3-VL-32B} & 32B & 44
    & 49.8 & 59.5 & 60.2 & 46.9 & 66.9 & 55.7 & \ovplain{56.45} \\
\oursrowshade{\textwidth}\ourslabel & 32B & $\leq44$
    & 55.4 & 62.4 & 56.4 & 46.8 & \textbf{76.0} & 54.2 & \ovgain{58.18}{1.73} \\
\addlinespace[1pt]
\textit{Qwen3.6-27B} & 27B & 44
    & 51.8 & \textbf{68.2} & \textbf{66.0} & 40.4 & 72.8 & 62.7 & \ovplain{60.80} \\
\oursrowshade{\textwidth}\ourslabel & 27B & $\leq44$
    & \textbf{60.7} & \underline{67.4} & 59.4 & 49.1 & 71.8 & \underline{65.2}
    & \ovgain{\textbf{62.51}}{1.70} \\
\addlinespace[1pt]
\textit{Qwen3.5-397B-A17B} & 397B-A17B-FP8 & 44
    & 54.2 & 62.4 & \underline{61.0} & 37.3 & 71.7 & 63.5 & \ovplain{58.84} \\
\oursrowshade{\textwidth}\ourslabel & 397B-A17B-FP8 & $\leq44$
    & \underline{59.4} & 64.4 & \underline{61.0} & \underline{49.3} & 70.1 & \textbf{65.6}
    & \ovgain{\underline{61.86}}{3.02} \\
\bottomrule
\end{tabular*}}
\end{table}

The unchanged configuration improves four of five MLLMs in
Table~\ref{tab:ovbench-main}; Qwen3-VL-8B changes by -0.04 points.
Qwen3.6-27B reaches the highest overall score, 62.51, and Qwen3.5-9B has
the largest paired increase, 3.05 points. For Qwen3.5-9B, Spatial Perception
and Temporal Hallucination Verification gain 8.4 and 7.0 points, respectively,
while Past Memory drops by 6.6 points. FP, THV, PM, SP, STP, and TP denote
Future Prediction, Temporal Hallucination Verification, Past Memory, Spatial
Perception, Spatio-Temporal Perception, and Temporal Perception. The full
16-subtask comparison is in Appendix~\ref{sec:ovbench-detailed}.

Across the three benchmarks, the same configuration improves all five
model-matched bases on OVO-Bench and StreamingBench Real-time and four on
OVBench.  The gains extend from the 8B model to the 397B mixture-of-experts
model, showing that the evidence-curation strategy remains effective across
the evaluated model scales and architectures.

\subsection{Ablation studies}

\paragraph{Component contributions.}
We use Qwen3.5-9B on complete OVO-Bench.  Table~\ref{tab:component-ablation}
starts from the full method and removes one
component at a time.  Without the look-back, the macro score falls
from 64.19 to 60.59, including an 8.36-point drop on Forward questions.
Removing historical retrieval lowers the macro score to 63.18 and Forward
accuracy by 4.09 points, showing that retrieval adds useful evidence even
with glance and look-back present. Removing glance curation lowers Avg.
to 63.26. Coverage aggregation raises Avg. from 63.29 to 64.19
while reducing returned historical frames per question from 6.76 to 4.13.
Appendix~\ref{sec:recent-curation-study} compares glance-curation rules;
Appendices~\ref{sec:additional-components}--\ref{sec:additional-subtasks}
further examine component replacements and task-specific contributions
with Qwen3.6-27B.

\begin{table}[t]
\caption{Component ablations on complete OVO-Bench (Qwen3.5-9B). Base uses 44 uniformly sampled frames. Each w/o row removes one component; $\Delta$ Avg. is relative to the full method.}
\label{tab:component-ablation}
\vspace{2pt}
\centering
{\fontsize{8.6}{9.8}\selectfont
\setlength{\tabcolsep}{2.2pt}
\renewcommand{\arraystretch}{1.02}
\begin{tabular*}{0.90\textwidth}{@{\extracolsep{\fill}}lccccc@{}}
\toprule
Setting & Backward & Real-time & Forward & Avg. & $\Delta$ Avg. \\
\midrule
Base & 51.37 & 71.66 & 54.34 & 59.12 & \textcolor{tableloss}{-5.07} \\
\midrule
\oursrowshade{0.90\textwidth}\methodcolor{} (full) & 60.33 & 76.43 & 55.83 & 64.19 & -- \\
\hspace*{0.9em}w/o look-back & 56.59 & 77.70 & 47.47 & 60.59 & \textcolor{tableloss}{-3.60} \\
\hspace*{0.9em}w/o glance curation & 60.91 & 75.90 & 52.98 & 63.26 & \textcolor{tableloss}{-0.93} \\
\hspace*{0.9em}w/o historical retrieval & 59.81 & 77.98 & 51.74 & 63.18 & \textcolor{tableloss}{-1.01} \\
\hspace*{0.9em}w/o coverage aggregation & 60.96 & 75.79 & 53.12 & 63.29 & \textcolor{tableloss}{-0.90} \\
\bottomrule
\end{tabular*}}
\end{table}

\begingroup
\setlength{\intextsep}{6pt}
\begin{table}[H]
\caption{Temporal coverage on complete OVO-Bench (Qwen3.5-9B, \%). Glance
and look-back give frame counts; -- means none. $\Delta$ Avg. is relative to
each block's reference row.}
\label{tab:temporal-coverage}
\vspace{1pt}
\centering
{\fontsize{8.3}{9.3}\selectfont
\setlength{\tabcolsep}{2.0pt}
\renewcommand{\arraystretch}{0.96}
\begin{tabular*}{0.86\textwidth}{@{\extracolsep{\fill}}cc!{\hspace{4pt}\vrule width 0.35pt\hspace{4pt}}ccccc@{}}
\toprule
Glance & Look back & Backward & Real-time & Forward & Avg. & $\Delta$ Avg. \\
\midrule
\rowcolor{tablegroup}\multicolumn{7}{l}{\textit{Glance size}} \\
4 & -- & 55.87 & 76.26 & 50.39 & 60.84
    & \textcolor{tableloss}{-1.17} \\
8 & -- & 56.25 & 77.30 & 49.68 & 61.08
    & \textcolor{tableloss}{-0.93} \\
12 & -- & 57.94 & 77.30 & 50.79 & \textbf{62.01} & -- \\
16 & -- & 57.55 & 77.88 & 50.00 & 61.81
    & \textcolor{tableloss}{-0.20} \\
\midrule
\rowcolor{tablegroup}\multicolumn{7}{l}{\textit{Look-back coverage}} \\
12 & -- & 57.94 & 77.30 & 50.79 & 62.01 & -- \\
12 & 4 & 60.08 & 76.60 & 52.35 & 63.01
    & \textcolor{tablegain}{+1.00} \\
12 & 8 & 60.91 & 76.49 & 56.01 & \textbf{64.47}
    & \textcolor{tablegain}{+2.46} \\
12 & 12 & 61.68 & 76.39 & 54.90 & 64.32
    & \textcolor{tablegain}{+2.31} \\
\midrule
\rowcolor{tablegroup}\multicolumn{7}{l}{\textit{Matched current-view size}} \\
20 & -- & 59.67 & 77.82 & 50.45 & 62.65 & -- \\
12 & 8 & 60.12 & 77.82 & 55.13 & \textbf{64.36}
    & \textcolor{tablegain}{+1.71} \\
\bottomrule
\end{tabular*}}
\end{table}
\endgroup

\paragraph{Temporal placement.}
Table~\ref{tab:temporal-coverage} isolates placement from glance curation.
Adding eight look-back frames to a 12-frame glance raises Avg. from 62.01 to
64.47.
With 20 frames in both views, the mixed layout still beats glance alone
(64.36 versus 62.65; video-cluster interval: $[0.31,3.15]$). Similar model
calls and image counts tie the gain to placement.

\paragraph{History capacity.}
Tables~\ref{tab:history-cap} and~\ref{tab:history-access} use SigLIP2 glance
curation. Table~\ref{tab:history-cap} varies the retrieved history cap.
Raising it from six to 24 frames improves Avg. from 63.59 to
64.60, with the largest change on Forward questions.

\begin{table}[t]
\caption{History-cap accuracy (\%) on complete OVO-Bench with Qwen3.5-9B. $\Delta$ Avg. uses six frames as reference.}
\label{tab:history-cap}
\vspace{2pt}
\centering
{\fontsize{9.0}{10.2}\selectfont
\setlength{\tabcolsep}{2.2pt}
\renewcommand{\arraystretch}{1.0}
\begin{tabular*}{0.76\textwidth}{@{\extracolsep{\fill}}lccccc@{}}
\toprule
History cap & Backward & Real-time & Forward & Avg. & $\Delta$ Avg. \\
\midrule
6 frames & 61.51 & 76.52 & 52.74 & 63.59 & -- \\
12 frames & 61.31 & 76.68 & 55.20 & 64.40 & \textcolor{tablegain}{+0.81} \\
18 frames & 61.43 & 76.62 & 55.21 & 64.42 & \textcolor{tablegain}{+0.83} \\
24 frames & 61.31 & 76.49 & 56.00 & \textbf{64.60} & \textcolor{tablegain}{+1.01} \\
\bottomrule
\end{tabular*}}
\end{table}

\FloatBarrier
\begin{table}[H]
\caption{History access on complete OVO-Bench with Qwen3.5-9B.
Scores are accuracy (\%); Calls and Images are per-question means.}
\label{tab:history-access}
\vspace{1pt}
\centering
{\fontsize{9.0}{10.2}\selectfont
\setlength{\tabcolsep}{2.2pt}
\renewcommand{\arraystretch}{1.0}
\begin{tabular*}{\textwidth}{@{\extracolsep{\fill}}lcccccc@{}}
\toprule
\multirow{2}{*}{Setting} & \multicolumn{4}{c}{Accuracy (\%)} &
    \multicolumn{2}{c}{Per-question usage} \\
\cmidrule(lr){2-5}\cmidrule(lr){6-7}
& Backward & Real-time & Forward & Avg. & Calls & Images \\
\midrule
\rowcolor{tablegroup}\multicolumn{7}{l}{\textit{History availability and tool use}} \\
No history & 60.03 & 77.39 & 51.72 & 63.05 & 1.00 & 19.43 \\
No evidence reconsideration & 61.61 & 76.16 & 48.04 & 61.94
    & 1.06 & 21.05 \\
Adaptive retrieval & 62.12 & 76.42 & 55.83 & 64.79
    & 1.67 & 39.06 \\
Always retrieve & 62.08 & 78.35 & 55.51 & \textbf{65.31}
    & 2.15 & 53.50 \\
\midrule
\rowcolor{tablegroup}\multicolumn{7}{l}{\textit{Historical frame selection with fixed tool decisions}} \\
Uniform sampling & 61.29 & 76.22 & 52.94 & 63.48
    & 1.67 & 39.06 \\
Visual-key matching & 62.53 & 76.41 & 55.86 & \textbf{64.93}
    & 1.67 & 39.06 \\
\bottomrule
\end{tabular*}}
\end{table}

\paragraph{Retrieval and source-frame selection.}
Table~\ref{tab:history-access} shows that adaptive retrieval raises Avg. from
63.05 without history to 64.79. Without evidence reconsideration, history
calls fall from 36.4\% to 5.5\% of questions and Avg. to 61.94. With tool
decisions and image counts fixed, visual-key matching outperforms uniform
sampling (64.93 versus 63.48; paired video-cluster interval: $[0.69,2.12]$).

\paragraph{Retrieval cost.}
Adaptive retrieval reduces model calls from 2.15 to 1.67 and images from
53.50 to 39.06 per question relative to always retrieving, counting repeated
first-call context. Avg. changes from 65.31 to 64.79; the paired
video-cluster interval for this 0.52-point difference includes zero.
Under matched call and image counts, Qwen3.6-27B also
retains a 4.53-point advantage over uniform selection
(Appendix~\ref{sec:additional-selection}). Its accuracy--latency study
finds a 5.32-point gain over Uniform 44 for about 0.70 s of additional mean
query time (Appendix~\ref{sec:additional-efficiency}).

\section{Limitations and Conclusion}
\method{} improves the evidence supplied to a frozen MLLM but cannot overcome
its perceptual or reasoning limits; joint training may raise this ceiling.
Counting results motivate selection that better preserves repeated
occurrences. Retrieval also increases tail latency, and the current agent
permits only one retrieval round. A ReAct-style agent could refine retrieval
after inspecting returned frames while managing the added response time.

Overall, \method{} curates multi-timescale visual evidence through a glance, look-back,
and agent-gated history retrieval. Across five frozen MLLMs, it reaches 70.07
on OVO-Bench, 86.12 on StreamingBench Real-time, and 62.51 on OVBench.
Ablations support the complementary roles of current-view curation and
historical recall.

\clearpage
\section*{AI use statement}

In this work, we independently formulated the research question,
developed the method, designed and conducted the experiments, analyzed the
results, and selected and verified the references. We used generative AI tools
only for language polishing and LaTeX formatting. We reviewed all
AI-assisted edits for technical accuracy and consistency with the experimental
results and cited sources. We take full responsibility for the final content
of this work, including any text or artifacts prepared with AI assistance.

\bibliographystyle{iclr2027_conference}
\bibliography{references}

\clearpage
\appendix
\section{Protocol Details}
\label{sec:protocol-details}

The state machine permits an initial model request, at most one
evidence-obligation retry, at most one schema repair, one real retrieval, and a
final answer request.  The MLLM supplies the retrieval decision and
arguments; the validator only requests a well-formed reconsideration and never
selects evidence itself.  Valid references must exist in the frozen causal
index and stay at or before the query frontier.  The final set contains at most
20 current candidates before semantic curation and 24 historical images.
Missing source videos, protocol failures, empty retrievals, timeouts, and
unparseable answers remain in the official denominator.

\paragraph{History updates and requests.}
For an incoming candidate $x_t$, the duplicate gate compares its
low-resolution appearance with the latest retained representative, using
exposure-compensated luminance changes, chrominance changes, and their
changed-pixel ratios. The coverage-gate threshold is 0.85.
A matching candidate extends the current record's temporal coverage;
otherwise it creates a new record.
A compact spatial appearance descriptor resolves a narrow boundary region;
scene changes and query-time anchors are retained.  A history-tool request is
$u_i=(z_i,\delta_i,\eta_i,\sigma_i,k_i)$, where $z_i$ names the missing visual
information, $\delta_i$ its temporal relation to known observations,
$\eta_i$ the evidence type, $\sigma_i$ the search scope, and $k_i$ the frame
budget.

Features from the MLLM visual encoder provide recent affinity on the three
streaming benchmarks; the LongVideoBench transfer rows use SigLIP2 recent
affinity.  Recent features are mean-pooled from the final unmerged visual
tokens in FP32 before normalization.  SigLIP2 embeddings provide history
localization in both cases.
These features are cached independently of the question.  The history
cache is filtered by the query frontier before retrieval, and the recent cache
is used only for the prescribed recent candidates; it never expands the
history candidate set.  Frames are
sent as JPEG images at quality 90 and at most 151,200 pixels per image.
\method{} uses temperature 0, seed 0, disabled thinking, and at most 256
output tokens. The Qwen3.6-27B uniform baseline protocols are distinguished
below.  LongVideoBench follows the official multiple-choice parser
with a deterministic query-seeded fallback; malformed outputs remain in the
denominator.

All paired StreamingBench references and \method{} runs use the same
single-letter answer constraint and the official first-character evaluator;
all 2,500 questions remain in the denominator.

\paragraph{Uniform-baseline protocols.}
The Qwen3.6-27B base in Table~\ref{tab:ovo-main} samples directly from native
frame indices in the observed prefix and answers the timestamped question
without an explicit system message. It serves as a direct uniform-video
question-answering reference. The studies in
Appendix~\ref{sec:additional-studies} instead use a shared causal candidate
pool and answering guidance to compare evidence-selection and inference
choices within the same evaluation framework. Both use all 3,035 OVO-Bench
questions, their original frontiers and labels, and the official scorer.

\begin{table}[!htbp]
\caption{Qwen3.6-27B uniform-baseline protocols on OVO-Bench. Both allow
up to 44 images and one answering call. Scores are reported within their
respective experiment groups.}
\label{tab:uniform-protocols}
\vspace{4pt}
\centering
{\small
\setlength{\tabcolsep}{4pt}
\renewcommand{\arraystretch}{1.12}
\begin{tabular}{@{}p{0.23\linewidth}p{0.34\linewidth}p{0.36\linewidth}@{}}
\toprule
Setting & Table~\ref{tab:ovo-main} base & Appendix~\ref{sec:additional-studies} Uniform 44 \\
\midrule
Sampling domain & Native observed-frame indices & Causal 1-fps candidates plus the exact query frontier \\
Frame selection & Uniform indices, rounded to frames & Uniform positions in the candidate pool \\
Short prefixes & Original even-frame-count rule; very short prefixes may repeat a frame & All available distinct candidates; no padding \\
Answer guidance & Timestamps and original question; no explicit system message & Shared answering guidance with retrieval disabled \\
Output-token cap & 128 & 256 \\
Mean images/question & 43.48 & 38.82 \\
Base Avg. (\%) & 60.94 & 64.09 \\
\method{} Avg. (\%) & 69.32 & 69.41 \\
Paired gain (points) & +8.38 & +5.32 \\
\bottomrule
\end{tabular}}
\end{table}

Both baselines use the same declared Qwen3.6-27B model identifier,
temperature 0, seed 0, disabled thinking, JPEG quality 90, and a
151,200-pixel limit per image. Table~\ref{tab:uniform-protocols} records
the input and prompting differences; gains are computed against the base
in the same experiment group. The frame-budget sweep varies only the
number of frames within the candidate-pool protocol.

\paragraph{StreamingBench task labels.}
OP, CR, CS, ATP, EU, TR, PR, SU, ACP, and CT denote Object Recognition,
Causal Reasoning, Clips Summarize, Attribute Recognition, Event Understanding,
Text-Rich Understanding, Prospective Reasoning, Spatial Understanding,
Action Recognition, and Counting, respectively.

\FloatBarrier
\section{Glance Curation}
\label{sec:recent-curation-study}

Table~\ref{tab:recent-curation} compares frame-selection rules within the same 12-frame glance and eight-frame look-back layout on complete OVO-Bench. The model-matched ViT used in the main method improves Avg. from 63.26 without curation to 64.19, a 0.93-point gain. Its score remains close to SigLIP2 at 64.60, while retaining 16.68 current frames per question on average instead of 19.43. This choice lets the MLLM's visual encoder curate the frames it subsequently reasons over.

\begin{table}[!h]
\caption{Glance-curation accuracy (\%) on complete OVO-Bench with Qwen3.5-9B. $\Delta$ Avg. is measured against no curation; bold and underline mark the highest and second-highest accuracy in each column.}
\label{tab:recent-curation}
\vspace{4pt}
\centering
{\fontsize{9.0}{10.2}\selectfont
\setlength{\tabcolsep}{2.4pt}
\renewcommand{\arraystretch}{1.05}
\begin{tabular*}{0.80\textwidth}{@{\extracolsep{\fill}}lccccc@{}}
\toprule
Curation & Backward & Real-time & Forward & Avg. & $\Delta$ Avg. \\
\midrule
None & \underline{60.91} & 75.90 & 52.98 & 63.26 & -- \\
Pixel similarity & 60.10 & 75.14 & 52.70 & 62.65 & \textcolor{tableloss}{-0.61} \\
SigLIP2 & \textbf{61.31} & \textbf{76.49} & \textbf{56.00} & \textbf{64.60} & \textcolor{tablegain}{+1.34} \\
Model-matched ViT & 60.33 & \underline{76.43} & \underline{55.83} & \underline{64.19} & \textcolor{tablegain}{+0.93} \\
\bottomrule
\end{tabular*}}
\end{table}

\FloatBarrier
\section{LongVideoBench Transfer with SigLIP2 Recent Curation}
\label{sec:lvb-siglip-transfer}

LongVideoBench tests offline long-video reasoning rather than a streaming
query frontier. We evaluate a SigLIP2 glance-curation variant with the same
context layout and history retrieval as the main method. Qwen3.6-27B obtains
the strongest result among these five variants at 59.24, followed by
Qwen3.5-397B-A17B at 58.79. Table~\ref{tab:lvb-main} reports this variant;
the streaming results use each MLLM's own visual encoder for glance curation.

\begin{table}[t]
\caption{LongVideoBench visual-only validation accuracy (\%)
\citep{wu2024longvideobench}. The \method{} rows use the separate SigLIP2
recent-curation variant; bold and underlined scores mark the best and second
best among the reported methods.}
\label{tab:lvb-main}
\vspace{4pt}
\centering
{\fontsize{8.2}{9.1}\selectfont
\setlength{\tabcolsep}{2.0pt}
\renewcommand{\arraystretch}{0.98}
\begin{tabular*}{0.82\textwidth}{@{\extracolsep{\fill}}lccc@{}}
\toprule
Method & Params & \#Frames & Accuracy \\
\midrule
\rowcolor{tablegroup}\multicolumn{4}{l}{\muted{\textit{Training-based methods}}} \\
\muted{StreamOPD\,\tabcite{wu2026streamopd}} & \muted{4B} & \muted{$\leq32$}
    & \muted{61.41} \\
\midrule
\rowcolor{tablegroup}\multicolumn{4}{l}{\textit{Off-the-shelf MLLMs}} \\
GPT-4o (0513)\,\tabcite{hurst2024gpt} & -- & 256 & \textbf{66.70} \\
Aria\,\tabcite{li2024aria} & 25B-A3.5B & 256 & \underline{64.20} \\
Gemini-1.5-Pro (0514)\,\tabcite{team2024gemini} & -- & 256 & 64.00 \\
LLaVA-Video-72B-Qwen2\,\tabcite{zhang2024llava} & 72B & 128 & 63.90 \\
Qwen2-VL-7B\,\tabcite{wang2024qwen2} & 7B & 256 & 55.60 \\
\midrule
\oursrowshade{0.82\textwidth}\methodcolor{} (\textit{Qwen3-VL-8B}) & 8B & $\leq44$ & 45.25 \\
\oursrowshade{0.82\textwidth}\methodcolor{} (\textit{Qwen3.5-9B}) & 9B & $\leq44$ & 57.29 \\
\oursrowshade{0.82\textwidth}\methodcolor{} (\textit{Qwen3-VL-32B}) & 32B & $\leq44$ & 55.35 \\
\oursrowshade{0.82\textwidth}\methodcolor{} (\textit{Qwen3.6-27B}) & 27B & $\leq44$ & 59.24 \\
\oursrowshade{0.82\textwidth}\methodcolor{} (\textit{Qwen3.5-397B-A17B}) & 397B-A17B-FP8 & $\leq44$ & 58.79 \\
\bottomrule
\end{tabular*}}
\end{table}

\FloatBarrier
\section{Detailed OVO-Bench Results}
\label{sec:detailed-results}

\begin{table}[!h]
\caption{Task-level OVO-Bench accuracy (\%). \textsc{SW} denotes
$+\method{}$ with at most 44 frames. The paired references use uniform 44-frame sampling. Bold and underline mark the highest and second-highest score for each task; ties share a mark.}
\label{tab:ovo-task-breakdown}
\vspace{4pt}
\centering
{\fontsize{7.0}{8.0}\selectfont
\setlength{\tabcolsep}{1.0pt}
\renewcommand{\arraystretch}{0.98}
\begin{tabular*}{\textwidth}{@{\extracolsep{\fill}}l*{10}{c}@{}}
\toprule
& \multicolumn{2}{c}{\textit{Qwen3-VL-8B}}
& \multicolumn{2}{c}{\textit{Qwen3.5-9B}}
& \multicolumn{2}{c}{\textit{Qwen3-VL-32B}}
& \multicolumn{2}{c}{\textit{Qwen3.6-27B}}
& \multicolumn{2}{c}{\textit{Qwen3.5-397B-A17B}} \\
\cmidrule(lr){2-3}\cmidrule(lr){4-5}\cmidrule(lr){6-7}\cmidrule(lr){8-9}\cmidrule(lr){10-11}
Task ($n$) & 44 & $+\textsc{SW}$ & 44 & $+\textsc{SW}$ & 44 & $+\textsc{SW}$
    & 44 & $+\textsc{SW}$ & 44 & $+\textsc{SW}$ \\
\midrule
\multicolumn{11}{l}{\textit{Backward}} \\
ASI (148) & 72.97 & 64.86 & 66.89 & 64.19 & 73.65
    & 70.27 & \underline{75.68} & 75.00 & \textbf{77.03} & \underline{75.68} \\
EPM (297) & 54.88 & 55.89 & 61.95 & 61.95 & 60.27
    & 59.60 & 65.66 & \textbf{68.69} & 64.98 & \underline{66.67} \\
HLD (186) & 27.96 & 55.91 & 25.27 & 54.84 & 48.39
    & \underline{59.14} & 33.33 & 53.23 & 40.86 & \textbf{73.66} \\
\addlinespace[1pt]
\multicolumn{11}{l}{\textit{Real-time}} \\
ACR (109) & 65.14 & 76.15 & 67.89 & 73.39 & 69.72
    & 77.06 & 71.56 & \textbf{83.49} & 70.64 & \underline{81.65} \\
ATR (116) & 72.41 & 78.45 & 72.41 & 78.45 & 80.17
    & \underline{81.90} & 80.17 & 81.03 & 79.31 & \textbf{85.34} \\
FPD (101) & 70.30 & 70.30 & 75.25 & 76.24 & 75.25
    & \textbf{78.22} & \underline{77.23} & \textbf{78.22} & \textbf{78.22} & 75.25 \\
OCR (149) & 78.52 & 88.59 & 81.88 & 89.26 & 83.22
    & 87.92 & 81.88 & \underline{91.95} & 80.54 & \textbf{92.62} \\
OJR (184) & 63.04 & 73.91 & 68.48 & 76.63 & 65.76
    & 73.91 & 71.74 & \underline{79.35} & 72.28 & \textbf{80.98} \\
STU (178) & 60.11 & 58.99 & 64.04 & 64.61 & 59.55
    & 65.73 & 66.29 & \underline{67.98} & 61.24 & \textbf{70.79} \\
\addlinespace[1pt]
\multicolumn{11}{l}{\textit{Forward}} \\
CRR (240) & 54.58 & 57.08 & 52.92 & 64.58 & 52.92
    & 59.17 & 46.25 & \textbf{68.33} & 60.83 & \underline{65.42} \\
REC (698) & 41.83 & 33.09 & 38.25 & 40.26 & 40.83
    & 32.81 & \textbf{48.85} & 44.70 & \underline{48.28} & 41.83 \\
SSR (629) & 67.09 & 70.91 & \underline{71.86} & 62.64 & 69.00
    & 63.43 & 54.21 & \textbf{72.97} & 70.59 & 64.07 \\
\bottomrule
\end{tabular*}}
\end{table}

\FloatBarrier

\section{Detailed OVBench Results}
\label{sec:ovbench-detailed}

The six task-family means in Table~\ref{tab:ovbench-main} are computed from
the subtasks below.  The official Avg. gives equal weight to each of the 16
subtasks, irrespective of family size.

\begin{table}[t]
\caption{OVBench accuracy by all 16 subtasks (\%). Columns follow the official six-task,
16-subtask taxonomy. Avg. is the unweighted mean of the 16 subtask
accuracies. Subtasks are shown to one decimal and Avg. to two. Methods without
a public task breakdown report only Avg.; colored subscripts in Avg. show
differences from the corresponding 44-frame reference. Bold and underlined
scores mark the best and second best. Method references are in
Table~\ref{tab:ovbench-main}.}
\label{tab:ovbench-detailed}
\vspace{4pt}
\centering
{\fontsize{6.5}{7.4}\selectfont
\setlength{\tabcolsep}{1.15pt}
\renewcommand{\arraystretch}{1.06}
\resizebox{\textwidth}{!}{%
\begin{tabular}{@{}lcc*{17}{c}@{}}
\toprule
Method & Params & \#Frames
    & \multicolumn{3}{c}{FP} & \multicolumn{3}{c}{THV}
    & \multicolumn{3}{c}{PM} & \multicolumn{2}{c}{SP}
    & \multicolumn{2}{c}{STP} & \multicolumn{3}{c}{TP} & \\
\cmidrule(lr){4-6}\cmidrule(lr){7-9}\cmidrule(lr){10-12}
\cmidrule(lr){13-14}\cmidrule(lr){15-16}\cmidrule(lr){17-19}
& & & AA & GSP & MP & AP & SV & OPr. & AR & PR & TR
    & AL & OPos. & AT & OT & AS & SL & OES & Avg. \\
\midrule
\rowcolor{tablegroup}\multicolumn{20}{l}{\muted{\textit{Training-based methods}}} \\
\muted{VideoChat-Online} & \muted{4B} & \muted{2 fps}
    & \muted{64.1} & \muted{59.7} & \muted{16.6} & \muted{63.1} & \muted{58.3} & \muted{62.8}
    & \muted{\textbf{42.2}} & \muted{54.4} & \muted{70.6} & \muted{\textbf{54.1}} & \muted{24.8} & \muted{88.7}
    & \muted{48.5} & \muted{73.0} & \muted{25.9} & \muted{71.7} & \muted{\ovplain{54.90}} \\
\muted{StreamForest} & \muted{7B} & \muted{1 fps}
    & \muted{--} & \muted{--} & \muted{--} & \muted{--} & \muted{--} & \muted{--}
    & \muted{--} & \muted{--} & \muted{--} & \muted{--} & \muted{--} & \muted{--}
    & \muted{--} & \muted{--} & \muted{--} & \muted{--} & \muted{\ovplain{60.50}} \\
\midrule
\rowcolor{tablegroup}\multicolumn{20}{l}{\textit{Off-the-shelf MLLMs}} \\
Gemini-1.5-Flash & -- & $\leq64$
    & \textbf{71.4} & 53.6 & 21.9 & 56.5 & 60.8 & 40.6
    & 36.7 & 47.9 & 62.5 & 32.3 & 37.5 & 87.0
    & 50.0 & \underline{83.3} & 22.3 & 46.9 & \ovplain{50.70} \\
LLaVA-OV & 7B & $\leq64$
    & 68.0 & 62.7 & \underline{35.9} & 58.4 & 50.3 & 46.5
    & 29.4 & 60.7 & 58.0 & 43.1 & 14.2 & 86.5
    & 49.7 & 70.7 & 28.1 & 30.2 & \ovplain{49.50} \\
Qwen2-VL & 7B & $\leq64$
    & 60.3 & 66.1 & 22.1 & 54.9 & 51.5 & 51.1
    & \underline{37.8} & 64.4 & 69.3 & 35.3 & 28.5 & \textbf{97.0}
    & 49.4 & 65.1 & 30.8 & 11.7 & \ovplain{49.70} \\
\midrule
\textit{Qwen3-VL} & 8B & 44
    & 57.7 & 66.7 & 26.1 & 53.2 & 59.8 & \underline{69.2}
    & 26.3 & 68.0 & 65.3 & 32.8 & 51.4 & 88.3
    & 54.7 & 74.4 & 26.9 & 61.5 & \ovplain{55.14} \\
\rowcolor{tableours}\ourslabel & 8B & $\leq44$
    & 65.4 & 74.1 & 22.7 & 56.7 & 59.4 & 63.7
    & 29.1 & 63.2 & 64.7 & 36.9 & 44.3 & 88.3
    & 52.5 & 75.3 & 27.7 & 57.8 & \ovloss{55.10}{0.04} \\
\addlinespace[0.35pt]
\textit{Qwen3.5} & 9B & 44
    & 53.8 & 70.5 & 29.4 & 46.4 & 62.2 & 58.2
    & 33.4 & 73.9 & 73.6 & 34.7 & 49.5 & 77.0
    & 51.6 & 79.5 & 26.8 & 62.5 & \ovplain{55.18} \\
\rowcolor{tableours}\ourslabel & 9B & $\leq44$
    & 67.9 & 75.4 & 25.8 & 57.9 & 61.2 & 68.6
    & 28.8 & 65.4 & 66.9 & 45.0 & 56.0 & 86.1
    & 48.8 & \textbf{83.3} & 30.9 & 63.7 & \ovgain{58.23}{3.05} \\
\addlinespace[0.35pt]
\textit{Qwen3-VL} & 32B & 44
    & 55.1 & 66.3 & 27.9 & 56.7 & 57.2 & 64.6
    & 29.7 & 73.4 & 77.6 & 31.3 & \textbf{62.5} & 85.7
    & 48.1 & 80.0 & 28.4 & 58.8 & \ovplain{56.45} \\
\rowcolor{tableours}\ourslabel & 32B & $\leq44$
    & 67.9 & 75.0 & 23.3 & 56.2 & 63.3 & 67.7
    & 30.3 & 71.4 & 67.5 & 36.6 & \underline{57.0} & \underline{94.3}
    & \underline{57.8} & 80.9 & 30.3 & 51.4 & \ovgain{58.18}{1.73} \\
\addlinespace[0.35pt]
\textit{Qwen3.6} & 27B & 44
    & 47.4 & 76.1 & 31.9 & 67.4 & \textbf{65.3} & \textbf{72.0}
    & 35.6 & \textbf{76.3} & \textbf{86.2} & 32.5 & 48.3 & 90.9
    & 54.7 & 78.1 & 29.8 & \underline{80.3} & \ovplain{60.80} \\
\rowcolor{tableours}\ourslabel & 27B & $\leq44$
    & 67.9 & \underline{78.4} & \underline{35.9} & \textbf{69.5} & \underline{64.8} & 68.0
    & 30.0 & 72.5 & 75.8 & \underline{46.9} & 51.4 & 89.6
    & 54.0 & 80.5 & \underline{35.1} & 80.0 & \ovgain{\textbf{62.51}}{1.70} \\
\addlinespace[0.35pt]
\textit{Qwen3.5} & {\fontsize{5.2}{6.0}\selectfont 397B-A17B-FP8} & 44
    & 53.8 & 71.9 & \textbf{36.8} & \underline{69.1} & 62.8 & 55.4
    & 25.9 & \underline{75.0} & \underline{82.2} & 26.3 & 48.3 & 83.5
    & \textbf{59.9} & 76.7 & 28.6 & \textbf{85.2} & \ovplain{58.84} \\
\rowcolor{tableours}\ourslabel & {\fontsize{5.2}{6.0}\selectfont 397B-A17B-FP8} & $\leq44$
    & \underline{69.2} & \textbf{78.9} & 30.1 & 64.4 & 64.4 & 64.3
    & 34.4 & 72.8 & 75.8 & 43.1 & 55.4 & 88.3
    & 51.9 & 82.3 & \textbf{35.2} & 79.4 & \ovgain{\underline{61.86}}{3.02} \\
\bottomrule
\end{tabular}}}
\end{table}

\FloatBarrier

\input{supplementary_studies.tex}

\FloatBarrier

\section{Qualitative Cases across Streaming Subtasks}
\label{sec:qualitative-subtasks}

\begin{figure}[H]
\centering
\includegraphics[page=1,width=\linewidth]{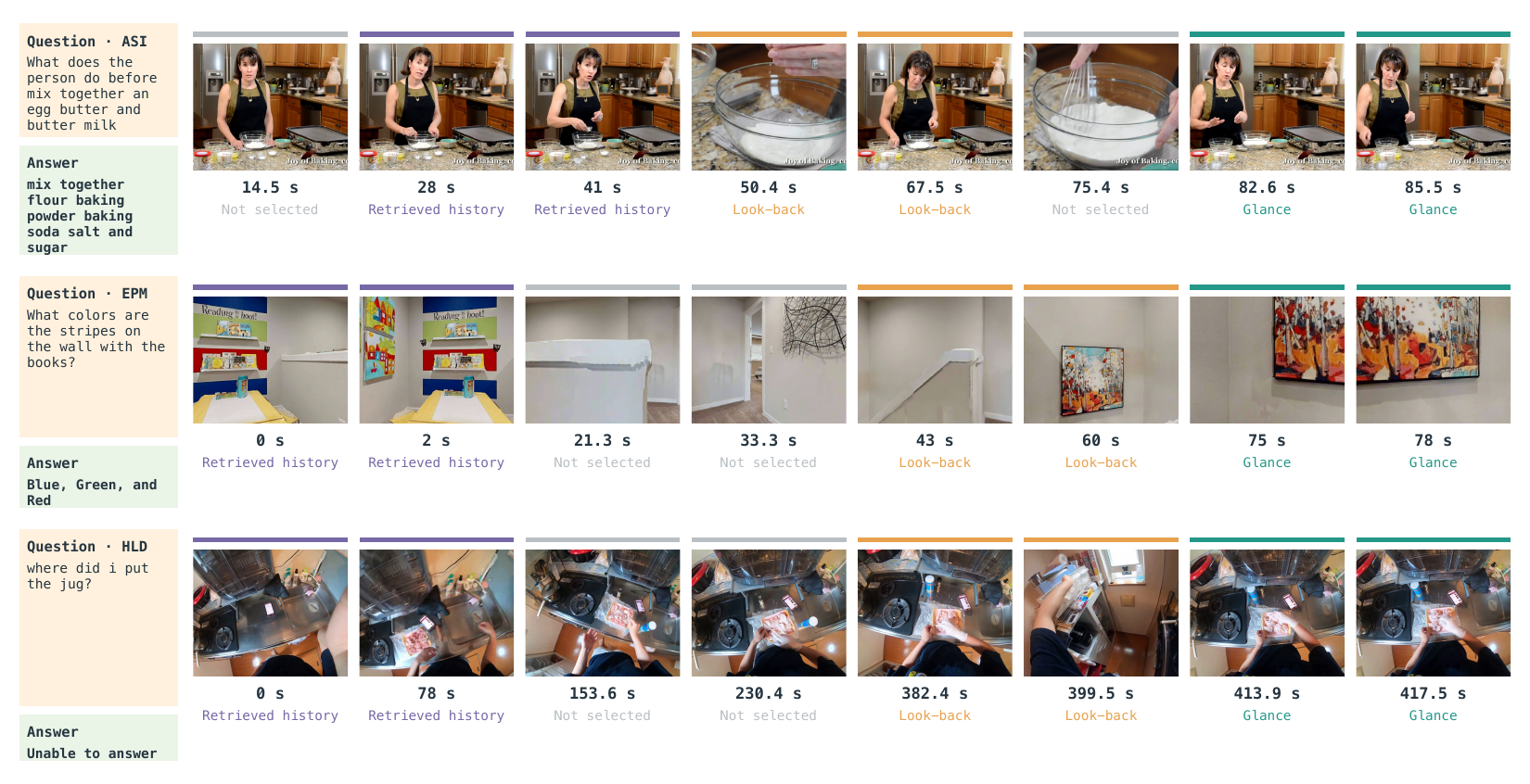}
\caption{Successful OVO-Bench Backward cases: ASI, EPM, and HLD, top to bottom. The EPM query recovers a brief book-wall detail from the start of the observed video.}
\label{fig:qual-ovo-backward}
\end{figure}

\begin{figure}[H]
\centering
\includegraphics[page=1,width=\linewidth]{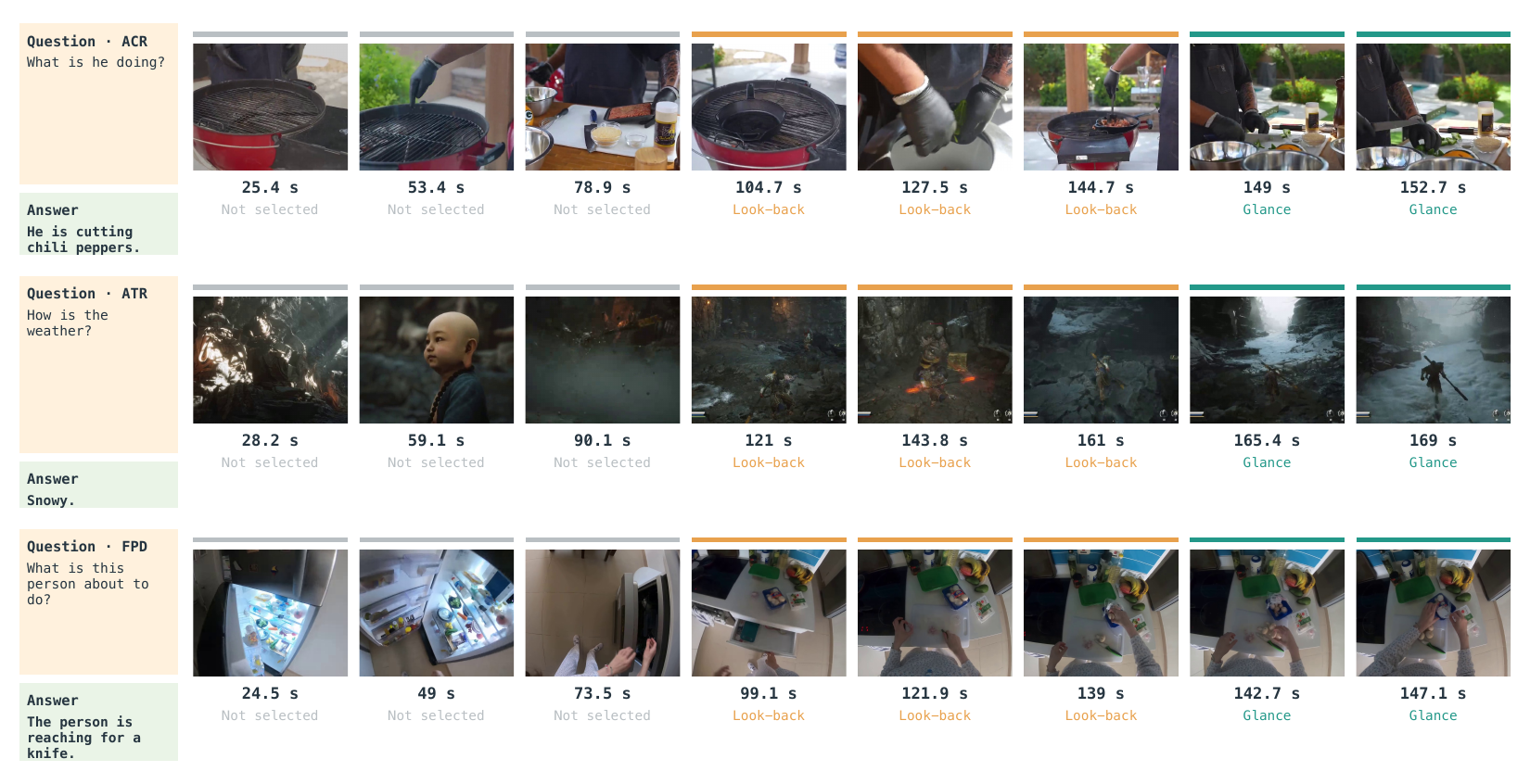}
\caption{Successful OVO-Bench Real-time cases: ACR, ATR, and FPD.}
\label{fig:qual-ovo-realtime-a}
\end{figure}

\begin{figure}[H]
\centering
\includegraphics[page=2,width=\linewidth]{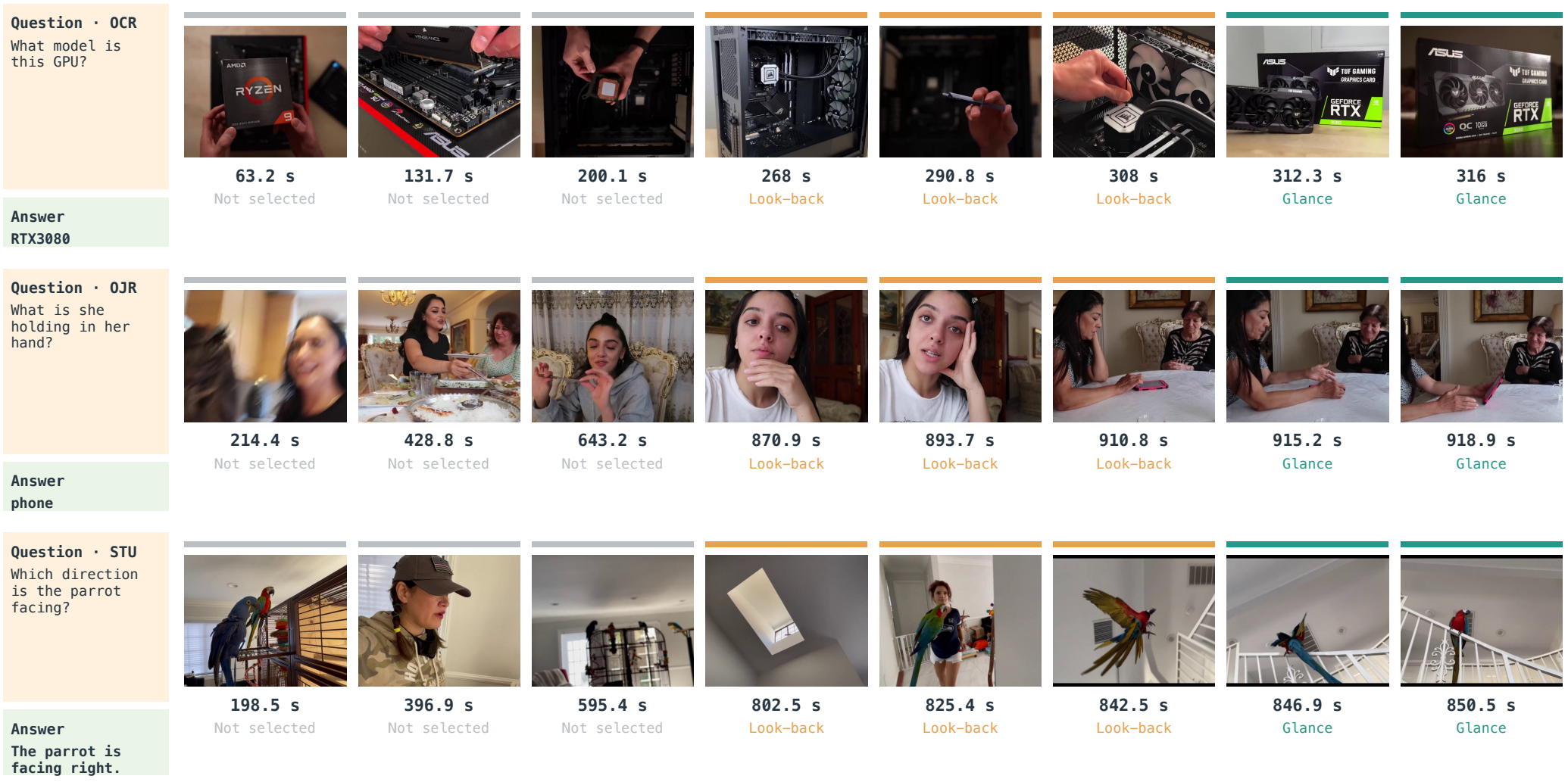}
\caption{Successful OVO-Bench Real-time cases: OCR, OJR, and STU.}
\label{fig:qual-ovo-realtime-b}
\end{figure}

\begin{figure}[H]
\centering
\includegraphics[page=1,width=\linewidth]{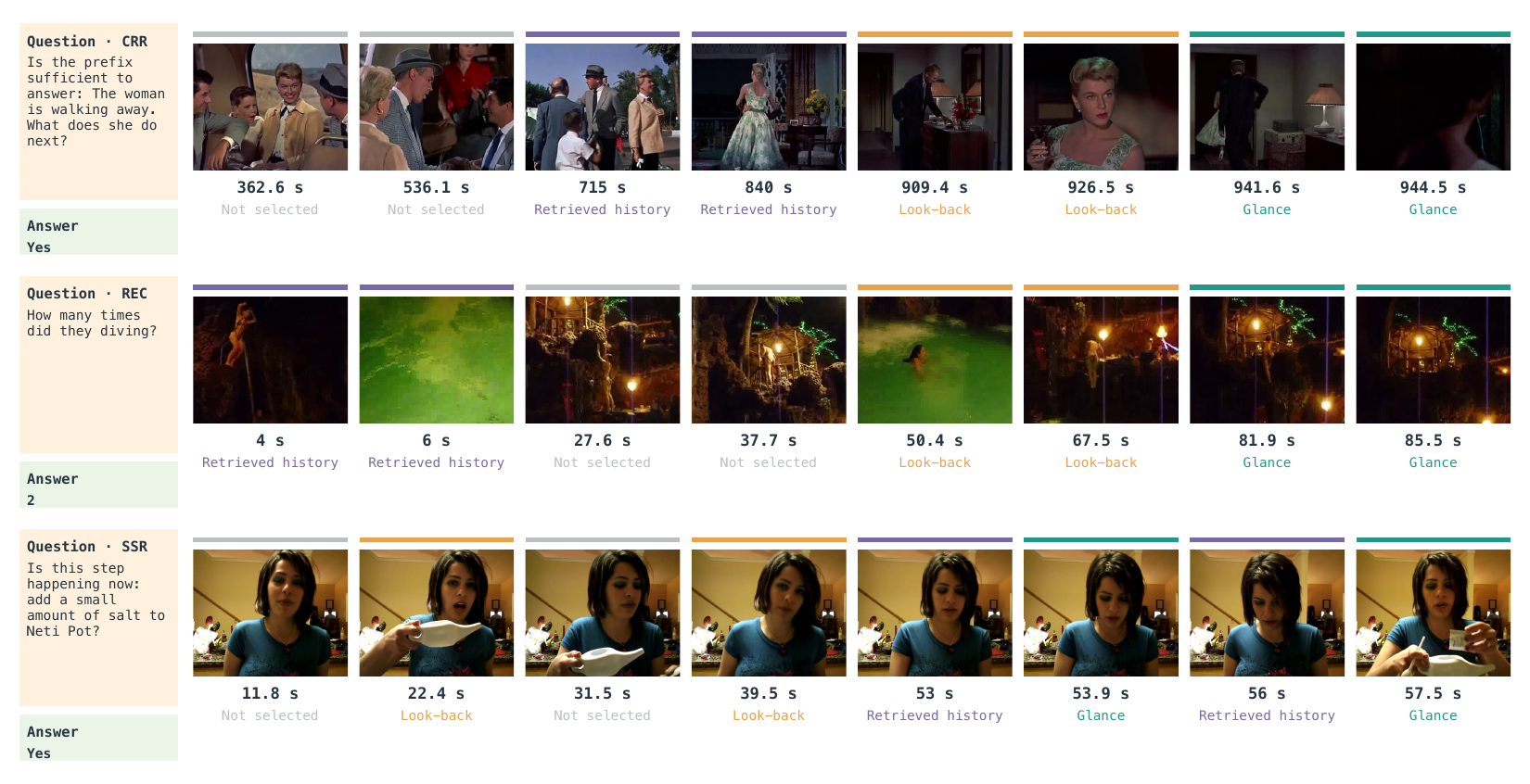}
\caption{Successful OVO-Bench Forward cases: CRR, REC, and SSR. CRR and SSR are yes/no judgments; REC asks for a cumulative count.}
\label{fig:qual-ovo-forward}
\end{figure}

\begin{figure}[H]
\centering
\includegraphics[page=1,width=\linewidth]{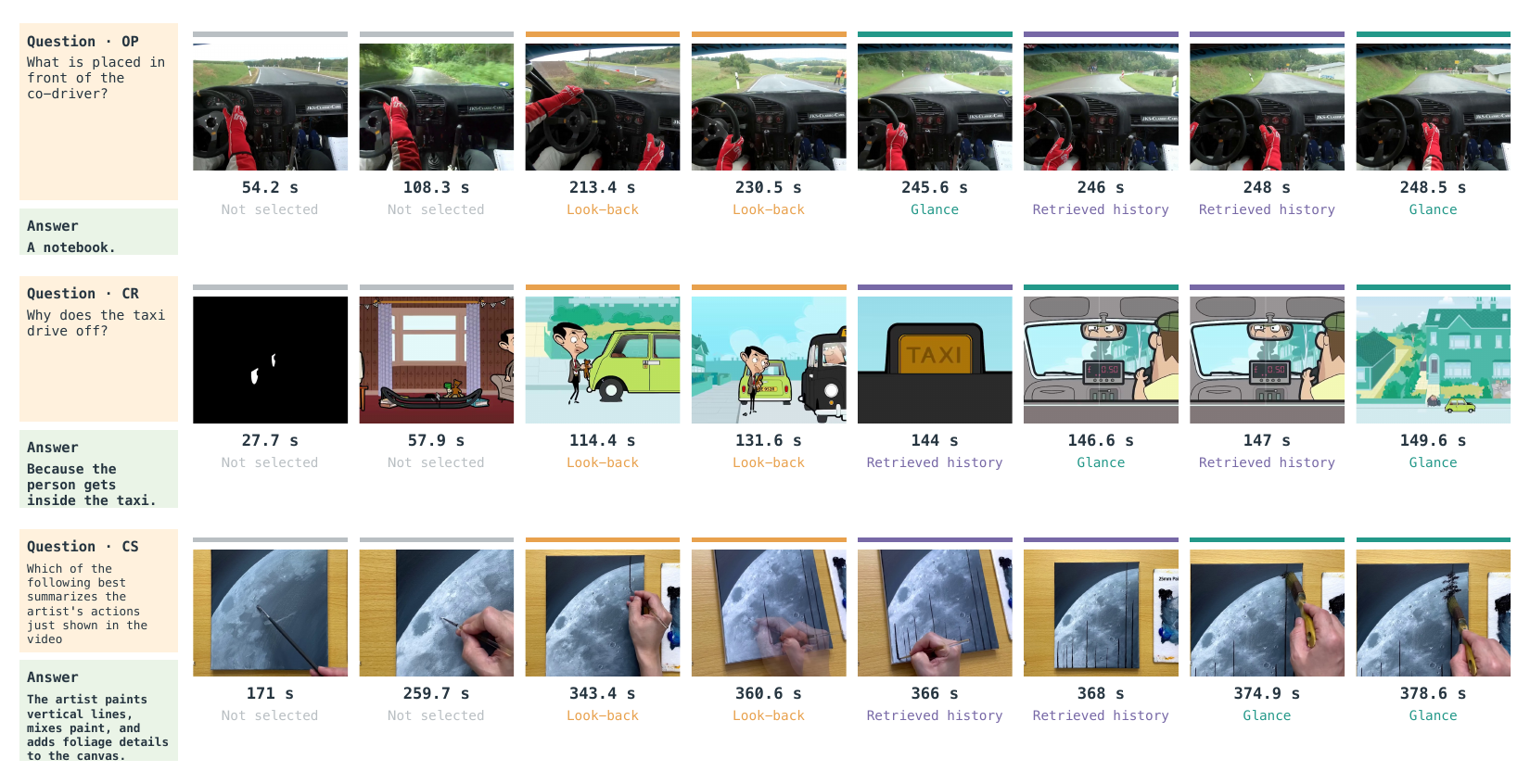}
\caption{Successful StreamingBench Real-time cases: OP, CR, and CS.}
\label{fig:qual-streaming-1a}
\end{figure}

\begin{figure}[H]
\centering
\includegraphics[page=2,width=\linewidth]{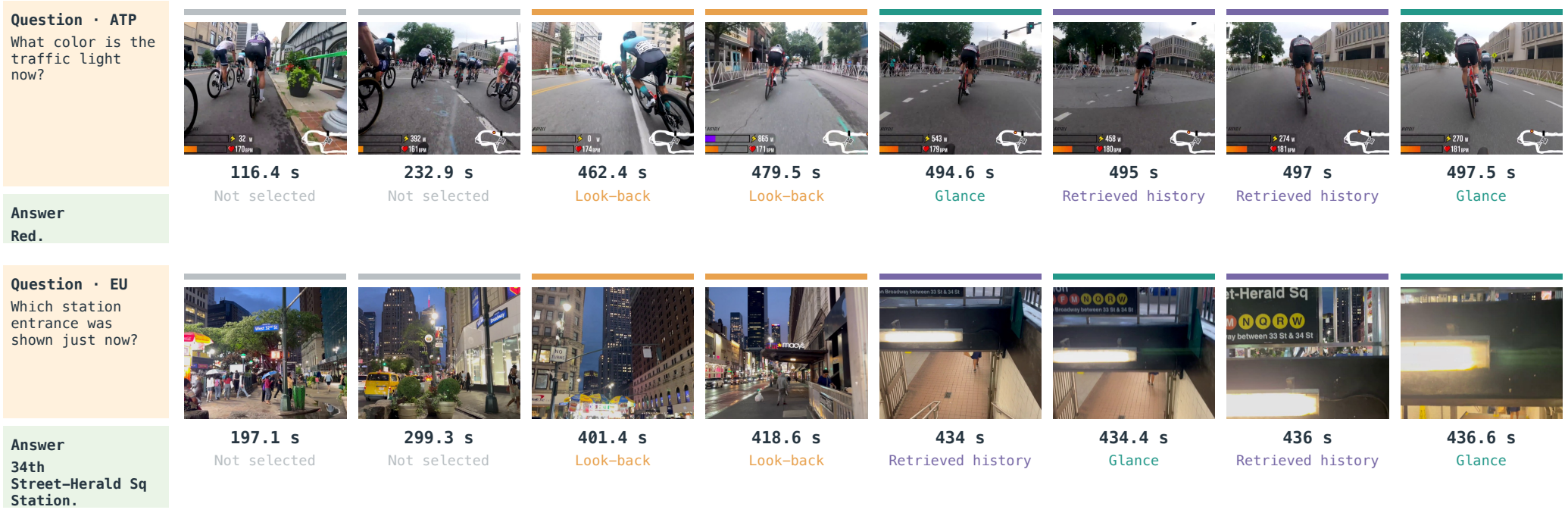}
\caption{Successful StreamingBench Real-time cases: ATP and EU.}
\label{fig:qual-streaming-1b}
\end{figure}

\begin{figure}[H]
\centering
\includegraphics[page=1,width=\linewidth]{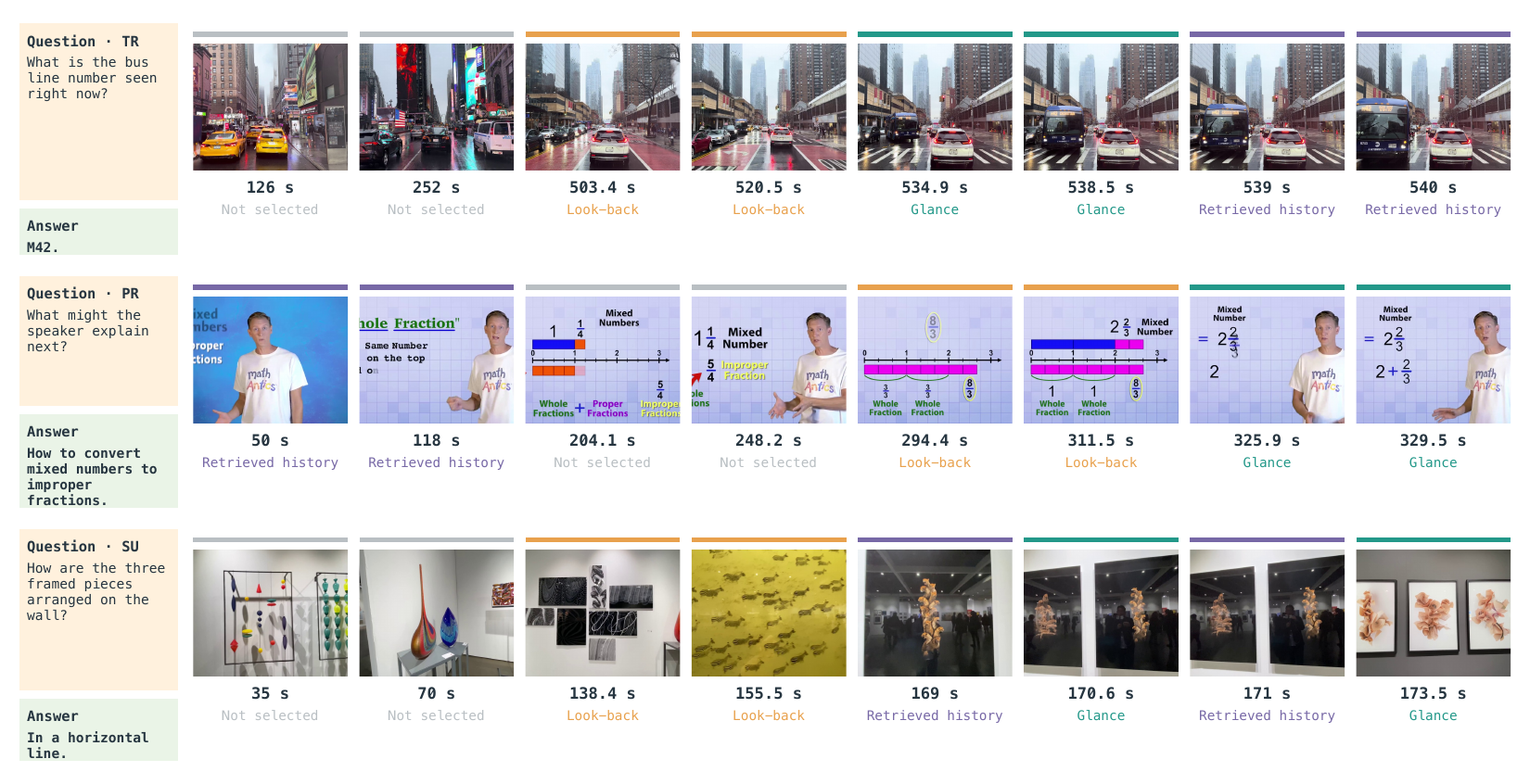}
\caption{Successful StreamingBench Real-time cases: TR, PR, and SU.}
\label{fig:qual-streaming-2a}
\end{figure}

\begin{figure}[H]
\centering
\includegraphics[page=2,width=\linewidth]{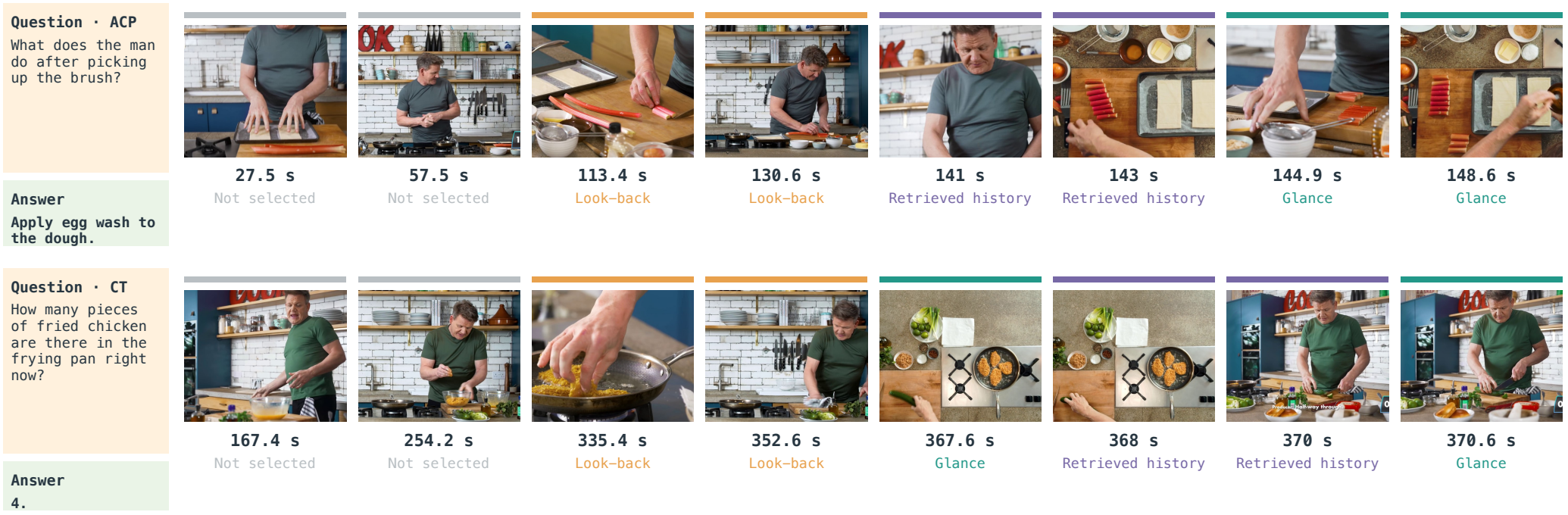}
\caption{Successful StreamingBench Real-time cases: ACP and CT.}
\label{fig:qual-streaming-2b}
\end{figure}

Figures~\ref{fig:qual-ovo-backward}--\ref{fig:qual-streaming-2b} show one correctly answered Qwen3.5-9B \method{} example for every OVO-Bench subtask and every StreamingBench Real-time task. Each row follows the observed video in time: purple frames were returned by history retrieval, orange frames belong to the uniformly sampled look-back, green frames to the recent glance, and gray frames were observed but not selected for that question. These labels come from the saved per-question evidence trace. No frame lies beyond the question frontier. The OVO-Bench HLD example correctly returns ``Unable to answer,'' as required by its annotation.

\raggedbottom

\end{document}

%% file: math_commands.tex
\usepackage{amsmath,amsfonts,bm}

\def\eqref#1{equation~\ref{#1}}

\def\1{\bm{1}}

\DeclareMathAlphabet{\mathsfit}{\encodingdefault}{\sfdefault}{m}{sl}
\SetMathAlphabet{\mathsfit}{bold}{\encodingdefault}{\sfdefault}{bx}{n}



%% file: supplementary_studies.tex
\section{Additional Component and Efficiency Studies}
\label{sec:additional-studies}

These supplementary experiments use frozen Qwen3.6-27B. OVO-Bench comparisons
cover all 3,035 questions from 644 source videos; the Counting study covers
all 193 questions from 55 videos in StreamingBench Real-time. Every method
uses only the prefix observed at the question frontier and the official
scorer. Uniform baselines use the candidate-pool protocol in
Table~\ref{tab:uniform-protocols}. We use common image preparation and
answer-generation settings within each comparison: at most 151,200 pixels per image, JPEG quality 90,
temperature 0, and a 256-token output limit per call. Failures remain in the
evaluation denominator. Reported confidence intervals use 10,000 paired
source-video bootstrap replicates.

\subsection{Comparing Corresponding Components}
\label{sec:additional-components}

Table~\ref{tab:component-ablation} tests component removal within \method{}.
Table~\ref{tab:additional-components} compares alternative mechanisms by
replacing one component at a time with its OASIS or VAM counterpart
\citep{liang2026oasis,li2026visual}, keeping the remaining components and
final answering interface fixed. The three comparisons concern history
access, memory construction, and context organization.

\paragraph{OASIS components.}
History access uses the same retained source-frame bank, retrieval requests,
and return budget. OASIS searches its event-description hierarchy and returns
source frames from the top two events; \method{} matches requests to visual
keys. Memory construction uses the same causal 1-fps candidates and
downstream visual ranking, replacing coverage records with OASIS event
construction, hierarchical merging, and retention rules, including accessible
child nodes and short-term buffers. Context organization shares the current
candidates and retrieved history, replacing glance/look-back packing with
OASIS NowWindow, short-term buffers, event evidence, and summaries.
History-access and construction comparisons hold retrieval requests fixed.

\paragraph{VAM components.}
VAM reports Gemini Embedding 2 for indexing and retrieval and Gemini 3 Flash
for answering. We instantiate its components with frozen
Qwen3-VL-Embedding-8B and the common frozen Qwen3.6-27B answering model.
History access uses VAM event records and retrieval on a shared source-frame
archive. Construction uses its adaptive writing and hierarchical retention
on the same causal 1-fps candidates, with downstream ranking fixed. Context
organization uses its event, spatial-memory, and temporal-context
representation within a shared pool of 20 current and at most 24 historical
candidates. Access and construction affect 795 historical-route questions;
the other 2,240 retain common predictions. Context organization covers all
3,035 questions. All comparisons are scored on the complete set, including
failures.

\begin{table}[!htbp]
\caption{Corresponding-component replacements on complete OVO-Bench with
Qwen3.6-27B (\%). Each replacement retains the other \method{} components.
Bold and underline mark the highest and second-highest score in each panel.}
\label{tab:additional-components}
\label{tab:additional-vam-components}
\vspace{4pt}
\centering
{\small
\setlength{\tabcolsep}{4pt}
\renewcommand{\arraystretch}{1.08}
\begin{tabular*}{\linewidth}{@{\extracolsep{\fill}}lcccc@{}}
\toprule
Configuration & Backward & Real-time & Forward & Avg. \\
\midrule
\rowcolor{tablegroup}\multicolumn{5}{l}{\textit{OASIS component replacements}} \\
History access & \textbf{65.92} & 78.32 & \underline{57.51} & \underline{67.25} \\
Memory construction & 63.21 & \underline{80.47} & 56.04 & 66.57 \\
Context organization & 63.84 & 80.02 & 56.12 & 66.66 \\
\oursrowshade{\linewidth}\methodcolor{} & \underline{65.80} & \textbf{80.68} & \textbf{61.76} & \textbf{69.41} \\
\midrule
\rowcolor{tablegroup}\multicolumn{5}{l}{\textit{VAM component replacements}} \\
History access & 62.78 & 79.43 & 55.66 & 65.96 \\
Memory construction & 62.95 & 79.52 & 55.13 & 65.87 \\
Context organization & \textbf{67.08} & \textbf{80.91} & \underline{56.41} & \underline{68.13} \\
\oursrowshade{\linewidth}\methodcolor{} & \underline{65.80} & \underline{80.68} & \textbf{61.76} & \textbf{69.41} \\
\bottomrule
\end{tabular*}}
\end{table}

\method{} exceeds the OASIS access, construction, and context replacements
by 2.16, 2.84, and 2.75 Avg. points, and the corresponding VAM replacements
by 3.45, 3.54, and 1.28 points. Forward gains are largest in both comparisons.
VAM context organization scores higher on Backward and Real-time, while
\method{} achieves a higher three-track average.

\subsection{Cumulative Contributions}
\label{sec:additional-cumulative}

Table~\ref{tab:additional-cumulative} starts from glance alone and adds the
remaining components. Glance alone gives the highest Real-time score in
this study, while look-back expands the temporal context and raises Backward
and Forward accuracy. Glance self-curation reduces current images from
19.51 to 16.68 per question while maintaining Avg. With these views already
present, historical retrieval and coverage aggregation raise Forward from
60.29 to 61.76. The full method reaches 69.41 Avg., compared with 65.23 for
glance alone. Appendix~\ref{sec:additional-subtasks} examines the task-level
changes underlying these track averages.

\begin{table}[!htbp]
\caption{Cumulative ablations on complete OVO-Bench with Qwen3.6-27B (\%).
Each row adds one component. History images/question includes questions
without retrieval; -- denotes zero. Bold and underline mark the highest and
second-highest accuracy in each column.}
\label{tab:additional-cumulative}
\vspace{4pt}
\centering
{\fontsize{8.8}{10.2}\selectfont
\setlength{\tabcolsep}{3pt}
\renewcommand{\arraystretch}{1.1}
\begin{tabular*}{\linewidth}{@{\extracolsep{\fill}}lccccc@{}}
\toprule
Configuration & Backward & Real-time & Forward & Avg. & \shortstack{History images\\/question} \\
\midrule
Glance & 61.77 & \textbf{82.99} & 50.95 & 65.23 & -- \\
+ look-back & 65.18 & 82.29 & 60.06 & 69.18 & -- \\
+ glance self-curation & 64.81 & \underline{82.59} & 60.29 & 69.23 & -- \\
+ frame-level history retrieval & \textbf{66.49} & 80.58 & \underline{60.92} & \underline{69.33} & 5.005 \\
\oursrowshade{\linewidth}+ coverage aggregation & \underline{65.80} & 80.68 & \textbf{61.76} & \textbf{69.41} & 3.090 \\
\bottomrule
\end{tabular*}}
\end{table}

Coverage aggregation reduces returned historical images by 38.26\%
(5.005 to 3.090 per question), with Avg. changing from 69.33 to 69.41.
The current input messages are identical with and without coverage aggregation. Across all logical
model calls, cumulative image inputs decrease from 30.401 to 28.444 per
question. The benefit is a more compact history contribution while preserving
answering accuracy.

\subsection{Task-level Contributions}
\label{sec:additional-subtasks}

Table~\ref{tab:additional-subtasks} reports every OVO-Bench subtask for
glance, glance with look-back, the curated current view, and the full method.

\begin{table}[H]
\caption{Task-level cumulative results on complete OVO-Bench with
Qwen3.6-27B (\%). Columns add look-back and self-curation; Full also includes
history retrieval and coverage. Bold and underline mark the highest and
second-highest score per row, with ties sharing a mark.}
\label{tab:additional-subtasks}
\vspace{4pt}
\centering
{\small
\setlength{\tabcolsep}{4pt}
\renewcommand{\arraystretch}{1.08}
\begin{tabular*}{0.96\linewidth}{@{\extracolsep{\fill}}lccccc@{}}
\toprule
Task & Questions & Glance & + look-back & + self-curation & Full \\
\midrule
\rowcolor{tablegroup}\multicolumn{6}{l}{\textit{Backward}} \\
EPM & 297 & 61.28 & \textbf{69.70} & \underline{68.35} & \underline{68.35} \\
ASI & 148 & 67.57 & \textbf{79.05} & \underline{77.70} & 76.35 \\
HLD & 186 & \textbf{56.45} & 46.77 & 48.39 & \underline{52.69} \\
\rowcolor{tablegroup}\multicolumn{6}{l}{\textit{Real-time}} \\
STU & 178 & \underline{74.16} & 71.35 & \textbf{74.72} & 69.10 \\
OJR & 184 & \textbf{83.15} & \underline{81.52} & 80.43 & 79.35 \\
ATR & 116 & \underline{82.76} & \textbf{83.62} & 81.90 & 81.03 \\
FPD & 101 & 77.23 & \underline{78.22} & \textbf{79.21} & \underline{78.22} \\
ACR & 109 & \textbf{85.32} & \underline{84.40} & \textbf{85.32} & \underline{84.40} \\
OCR & 149 & \textbf{95.30} & \underline{94.63} & 93.96 & 91.95 \\
\rowcolor{tablegroup}\multicolumn{6}{l}{\textit{Forward}} \\
REC & 698 & 25.79 & 38.11 & \underline{38.68} & \textbf{45.13} \\
CRR & 240 & 52.50 & 66.25 & \underline{66.67} & \textbf{67.50} \\
SSR & 629 & 74.56 & \textbf{75.83} & \underline{75.52} & 72.66 \\
\midrule
Avg. & 3,035 & 65.23 & 69.18 & \underline{69.23} & \textbf{69.41} \\
\bottomrule
\end{tabular*}}
\end{table}

Adding look-back improves ASI by 11.49 points and EPM by 8.42 points,
providing observations from the preceding activity. Glance self-curation
then improves STU by 3.37 points while reducing current-frame inputs.
Adding history retrieval with coverage to the curated current view raises
REC by 6.45 points, HLD by 4.30 points, and CRR by 0.83 points.
REC counts repeated events observed so far; HLD tests whether a question can
be answered from the available evidence; CRR tests whether sufficient clues
have appeared to answer. These gains show how recovered evidence supports
different demands beyond the current view. Real-time accuracy decreases
from 82.59 to 80.68 with the historical contribution, while the full method
achieves the highest three-track average in the cumulative study.

\subsection{Frame Selection and Inference Budget}
\label{sec:additional-selection}

We apply BOLT \citep{liu2025bolt}, AKS \citep{tang2025adaptive}, and HiMu
\citep{ben2026himu} to the same observed prefixes with Qwen3.6-27B as the
answering model. BOLT and AKS retain their original CLIP-L/14 and CLIP-B/32
selection scorers, respectively. HiMu retains its planner and visual experts
in a visual-prefix adaptation. Table~\ref{tab:additional-selection} evaluates
these choices under streaming access constraints, rather than access to a
completed video. \method{} improves Avg. by 5.32 points over uniform sampling
and by 7.58--8.47 points over these selectors, with higher scores on all
three tracks.

\begin{table}[!htbp]
\caption{Frame-selection comparison on complete OVO-Bench with Qwen3.6-27B
(\%). Every method uses the same observed prefix. HiMu is adapted to the
visual prefix. Bold and underline mark the highest and second-highest
accuracy in each column.}
\label{tab:additional-selection}
\vspace{4pt}
\centering
{\small
\setlength{\tabcolsep}{4pt}
\renewcommand{\arraystretch}{1.08}
\begin{tabular*}{0.94\linewidth}{@{\extracolsep{\fill}}lcccc@{}}
\toprule
Method & Backward & Real-time & Forward & Avg. \\
\midrule
Uniform 44 & \underline{60.46} & \underline{73.78} & \underline{58.02} & \underline{64.09} \\
BOLT & 57.23 & 69.64 & 56.97 & 61.28 \\
AKS & 57.27 & 70.74 & 54.82 & 60.94 \\
HiMu (visual-prefix adaptation) & 56.03 & 72.40 & 57.04 & 61.83 \\
\oursrowshade{0.94\linewidth}\methodcolor{} & \textbf{65.80} & \textbf{80.68} & \textbf{61.76} & \textbf{69.41} \\
\bottomrule
\end{tabular*}}
\end{table}

We compare evidence-selection strategies under two inference budgets
(Table~\ref{tab:additional-budget}). The first follows \method{}'s
per-question call schedule and image counts, but selects frames uniformly
and generates its own intermediate reasoning and provisional answers. The
second always uses two calls, with up to 20 and 44 images, respectively.
Both use the same 256-token output limit per call. Image counts sum all
logical model requests, including repeated context; input and output tokens
are actual reported usage, not matched totals. Missing usage affects fewer
than 0.31\% of requests.

\begin{table}[!htbp]
\caption{Inference-budget comparison on complete OVO-Bench with Qwen3.6-27B.
Usage is averaged per question across all calls. Bold and underline mark the
highest and second-highest Avg.; token totals use available usage records.}
\label{tab:additional-budget}
\vspace{4pt}
\centering
{\fontsize{8.8}{10.2}\selectfont
\setlength{\tabcolsep}{3pt}
\renewcommand{\arraystretch}{1.1}
\begin{tabular*}{\linewidth}{@{\extracolsep{\fill}}lccccc@{}}
\toprule
Configuration & Avg. (\%) & Calls & Images & \shortstack{Input\\tokens} & \shortstack{Output\\tokens} \\
\midrule
Uniform 44 & 64.09 & 1.000 & 38.823 & 6,448.7 & 2.0 \\
Uniform: matched calls/images & \underline{64.88} & 1.525 & 28.444 & 4,934.4 & 109.2 \\
Uniform $20\!\to\!44$ & 63.54 & 2.000 & 57.954 & 9,948.6 & 4.0 \\
\oursrowshade{\linewidth}\methodcolor{} & \textbf{69.41} & 1.525 & 28.444 & 6,612.5 & 30.1 \\
\bottomrule
\end{tabular*}}
\end{table}

\method{} retains a 4.53-point advantage with calls and images matched.
Always using two calls and more image and input-token capacity also does not
recover this gain: Uniform $20\!\to\!44$ scores 63.54, 5.87 points below
\method{}. Together, these comparisons support the value of evidence
selection and organization beyond the number of calls or images supplied.

\subsection{Query Latency and Shared Stream Processing}
\label{sec:additional-efficiency}

We evaluate Uniform 12, 20, 32, and 44, \method{}, and HiMu under the same
candidate-pool and answering protocol. Accuracy uses all 3,035 OVO-Bench
questions. Latency uses 120 questions from 107 videos, sampled across all 12
tasks and stratified by observed-prefix length, with three repeats per
method. Methods are interleaved in a fixed randomized order. Mean, P50, and
P95 are weighted to reflect the full task and prefix-length distribution.
The answering model is Qwen3.6-27B in BF16 on two H800 80GB GPUs; HiMu's
visual experts run on an L20Y.

Timing covers query-dependent frame selection, context construction, text
encoding, retrieval, image preparation, planning, and all answering calls
and retries. Decoded RGB frames and question-independent image features
are cached; their reusable processing costs are reported separately below.
No question-dependent retrieval results or answers are cached.

\begin{table}[!htbp]
\caption{Accuracy--latency comparison with Qwen3.6-27B. Accuracy is measured
on complete OVO-Bench; latency (seconds) uses the same 120-question cohort
with three repeats per method. Bold and underline mark the highest and
second-highest accuracy, and the lowest and second-lowest latency.}
\label{tab:additional-latency}
\vspace{4pt}
\centering
{\small
\setlength{\tabcolsep}{4pt}
\renewcommand{\arraystretch}{1.08}
\begin{tabular*}{\linewidth}{@{\extracolsep{\fill}}lcccc@{}}
\toprule
Method & Avg. (\%) & Mean & P50 & P95 \\
\midrule
Uniform 12 & 63.14 & \textbf{1.43} & \textbf{1.40} & \textbf{2.24} \\
Uniform 20 & 63.18 & \underline{1.94} & \underline{1.98} & \underline{3.43} \\
Uniform 32 & 63.47 & 2.60 & 2.59 & 5.03 \\
Uniform 44 & \underline{64.09} & 3.22 & 3.21 & 6.29 \\
\oursrowshade{\linewidth}\methodcolor{} & \textbf{69.41} & 3.91 & 2.23 & 12.50 \\
HiMu (visual-prefix adaptation) & 61.83 & 15.89 & 12.72 & 40.70 \\
\bottomrule
\end{tabular*}}
\end{table}

Increasing the uniform budget from 12 to 44 frames raises accuracy by 0.95
points and mean latency by 1.79 s. Relative to Uniform 44, \method{} gains
5.32 points for about 0.70 s of additional mean query time. Its median is
lower (2.23 versus 3.21 s), while retrieval increases P95 to 12.50 s.
Compared with HiMu, it gains 7.58 points with substantially lower mean
query latency. The paired mean-time difference from Uniform 44 is 0.697 s
with a 95\% video-bootstrap interval of $[-0.141,1.578]$ s. These results
characterize the additional query cost of recovering more relevant evidence.
All 360 measurements per method are retained, including three HiMu timing
failures affecting one question; its successful-run mean is 15.95 s.

\paragraph{Additional selector timings.}
A separate cohort of 24 questions from 24 videos covers the same 12 tasks
and includes BOLT and AKS, with three repeats per method
(Table~\ref{tab:additional-latency-small}). It uses the same cache boundary
and includes query-time selection, retrieval, image preparation, and
answering calls. Comparisons in this table are made within this cohort.
Relative to always making two calls, \method{} reduces mean latency by
39\%; the paired 95\% interval for the mean difference is $[-4.49,-0.89]$ s.

\begin{table}[!htbp]
\caption{Additional query-latency measurements (seconds) on a separate
24-question cohort, with 72 runs per method. Bold and underline mark the
lowest and second-lowest latency in each column.}
\label{tab:additional-latency-small}
\vspace{4pt}
\centering
{\small
\setlength{\tabcolsep}{5pt}
\renewcommand{\arraystretch}{1.08}
\begin{tabular*}{0.86\linewidth}{@{\extracolsep{\fill}}lccc@{}}
\toprule
Method & Mean & P50 & P95 \\
\midrule
Uniform 44 & 4.55 & 4.28 & 7.96 \\
BOLT & 4.97 & 4.94 & 8.46 \\
AKS & \underline{4.06} & 3.92 & \underline{7.23} \\
Uniform $20\!\to\!44$ & 7.23 & 7.02 & 12.63 \\
\oursrowshade{0.86\linewidth}\methodcolor{} & 4.41 & \underline{3.21} & 12.38 \\
\method{} w/o retrieval & \textbf{3.05} & \textbf{2.82} & \textbf{5.69} \\
\bottomrule
\end{tabular*}}
\end{table}

We separately measure reusable stream processing on an L20Y for six videos,
covering 43.77 video-minutes and 51 questions
(Table~\ref{tab:additional-frontends}). The two visual frontends require a
combined 0.891 GPU-seconds per video-minute. Encoding uses CUDA-event timing;
peak allocated GPU memory and source-video decoding time are measured
separately for each frontend.

\begin{table}[!htbp]
\caption{Shared visual processing on an L20Y: six videos, 43.77
video-minutes, and 51 questions. Each frontend's memory peak and decoding
time are measured separately; the peaks do not represent joint residency.}
\label{tab:additional-frontends}
\vspace{4pt}
\centering
{\small
\setlength{\tabcolsep}{5pt}
\renewcommand{\arraystretch}{1.1}
\begin{tabular*}{\linewidth}{@{\extracolsep{\fill}}lccc@{}}
\toprule
Frontend & \shortstack{Encoding\\(GPU s/video-min)} & \shortstack{Peak GPU\\memory (GB)} & \shortstack{Decoding\\(s, total)} \\
\midrule
History encoder & 0.714 & 1.66 & 116.62 \\
Current-view encoder & 0.177 & 1.41 & 153.83 \\
\bottomrule
\end{tabular*}}
\end{table}

With image features available, incremental coverage-index updates take
0.0082 s over this sample. Coverage reduces index storage from 8.184 MB for
frame-level records to 1.176 MB, an 85.62\% reduction excluding the source
video archive. These costs are incurred during stream processing and reused
across questions; Table~\ref{tab:additional-latency} reports the subsequent
query costs.

\subsection{Counting and Evidence Selection}
\label{sec:additional-counting}

We examine the Counting decrease in Table~\ref{tab:streaming-main} by varying
glance curation, coverage aggregation, and historical retrieval on the full
Counting subset. Uniform and \method{} results in
Table~\ref{tab:additional-counting} use the shared candidate-pool and answering
protocol in Appendix~\ref{sec:protocol-details}; Table~\ref{tab:streaming-main}
uses the main evaluation protocol, with the same sampling and prompting
distinction as the OVO-Bench comparisons. The four curation/aggregation combinations all retain
history retrieval; disabling aggregation therefore uses frame-level history,
not a history-free context. A further w/o-retrieval run keeps the exact
current-view images used by the full setting for all 193 questions.

\begin{table}[!htbp]
\caption{StreamingBench Counting accuracy with Qwen3.6-27B: 193 questions
from 55 videos. $\checkmark$ denotes an enabled component. Bold and
underline mark the highest and second-highest accuracy in each block.}
\label{tab:additional-counting}
\vspace{4pt}
\centering
{\small
\setlength{\tabcolsep}{4pt}
\renewcommand{\arraystretch}{1.1}
\begin{tabular*}{\linewidth}{@{\extracolsep{\fill}}lccccc@{}}
\toprule
Method & \shortstack{Glance\\curation} & Coverage & Retrieval & Correct/total & Accuracy (\%) \\
\midrule
Uniform 44 & -- & -- & -- & 131/193 & \textbf{67.88} \\
\midrule
\method{} & -- & -- & $\checkmark$ & 100/193 & 51.81 \\
 & -- & $\checkmark$ & $\checkmark$ & 102/193 & \underline{52.85} \\
 & $\checkmark$ & -- & $\checkmark$ & 101/193 & 52.33 \\
\oursrowshade{\linewidth}\methodcolor{} (full) & $\checkmark$ & $\checkmark$ & $\checkmark$ & 104/193 & \textbf{53.89} \\
\method{} w/o retrieval & $\checkmark$ & -- & -- & 88/193 & 45.60 \\
\bottomrule
\end{tabular*}}
\end{table}

Historical retrieval increases accuracy from 45.60 to 53.89 with the same
current-view images, an 8.29-point gain with a paired 95\% interval of
$[3.94,12.86]$. The full setting also scores highest among the four
curation/aggregation combinations. Disabling these components does not close
the gap to uniform sampling, so the Counting decrease cannot be explained
simply by either curation or aggregation alone. Counting remains an area for
improvement: count-aware selection that preserves repeated occurrences
across the observed video is a promising direction.